%% file: main.tex
\documentclass[10pt,twocolumn,letterpaper]{article}
\usepackage[pagenumbers]{cvpr}

\usepackage{algorithm,algpseudocode,float}
\usepackage{lipsum}
\usepackage{booktabs}
\usepackage{amsmath}
\usepackage{multirow}
\usepackage{xcolor}
\usepackage[table]{xcolor}
\usepackage{amssymb}
\usepackage{pifont}
\usepackage{cuted}
\usepackage{caption}
\usepackage{newtxtext,newtxmath}

\definecolor{cvprblue}{rgb}{0.21,0.49,0.74}
\usepackage[pagebackref,breaklinks,colorlinks,allcolors=cvprblue]{hyperref}

\def\paperID{*****}
\def\confName{CVPR}
\def\confYear{2026}

\title{Fast3Dcache: Training-free 3D Geometry Synthesis Acceleration}
\author{
Mengyu Yang${^{1,2}}$ \quad Yanming Yang${^{1}}$ \quad Chenyi Xu${^{1}}$ \quad Chenxi Song${^{1}}$ \quad Yufan Zuo${^{1}}$ \quad Tong Zhao${^{1}}$ \\
Ruibo Li${^{3}}$ \quad Chi Zhang${^{1}}$\thanks{Corresponding author.} \\[0.5em] 
${^{1}}$AGI Lab, Westlake University \quad ${^{2}}$University of Electronic Science and Technology of China \\
${^{3}}$Nanyang Technological University
\\
{\tt\small Project Page: \href{https://fast3dcache-agi.github.io}{https://fast3dcache-agi.github.io}}
}

\begin{document}
\maketitle
\input{sec/0_abstract}    
\input{sec/1_intro}
\input{sec/2_related_work}
\input{sec/3_Preliminaries}
\input{sec/4_Method}
\input{sec/5_Experiment}
\input{sec/6_Conclusion}
{
    \small
    \bibliographystyle{ieeenat_fullname}
    \bibliography{main}
}
\input{sec/X_suppl}
\end{document}

%% file: sec/0_abstract.tex
\begin{abstract}
Diffusion models have achieved impressive generative quality across modalities like 2D images, videos, and 3D shapes, but their inference remains computationally expensive due to the iterative denoising process. While recent caching-based methods effectively reuse redundant computations to speed up 2D and video generation, directly applying these techniques to 3D diffusion models can severely disrupt geometric consistency. In 3D synthesis, even minor numerical errors in cached latent features accumulate, causing structural artifacts and topological inconsistencies. To overcome this limitation, we propose Fast3Dcache, a training-free geometry-aware caching framework that accelerates 3D diffusion inference while preserving geometric fidelity. Our method introduces a Predictive Caching Scheduler Constraint (PCSC) to dynamically determine cache quotas according to voxel stabilization patterns and a Spatiotemporal Stability Criterion (SSC) to select stable features for reuse based on velocity magnitude and acceleration criterion. Comprehensive experiments show that Fast3Dcache accelerates inference significantly, achieving up to a 27.12\% speed-up and a 54.83\% reduction in FLOPs, with minimal degradation in geometric quality as measured by Chamfer Distance (2.48\%) and F-Score (1.95\%).
\end{abstract}

%% file: sec/1_intro.tex
\section{Introduction}
\label{sec:intro}

Diffusion models and Flow matching have demonstrated remarkable success in generating high-fidelity content across various modalities including 2D images \cite{lipman2022flow,ho2020denoising,song2020denoising,liu2022flow}, videos \cite{wan2025wan,kong2024hunyuanvideo,blattmann2023stable}, and 3D assets \cite{wu2025direct3d,xiang2025structured,hunyuan3d22025tencent}. However, a significant drawback is their computationally intensive and slow inference process, which relies on a sequential and iterative denoising procedure.
To alleviate this computational bottleneck, recent studies have explored caching-based acceleration techniques that exploit redundancy in intermediate computations \cite{zou2024accelerating,zhou2025less,wu2025fast,liu2025plug,selvaraju2024fora,zhang2025blockdance,ma2024deepcache,liu2025reusing,lv2024fastercache,chen2025accelerating}. The core idea is to cache and reuse computations from previous timesteps, thereby reducing the need for repeated inference. These caching methods have shown considerable success in accelerating 2D image and video generation.

However, directly extending these caching strategies to 3D diffusion models presents significant challenges. 
In 2D or video generation, caching methods typically exploit redundancy in pixel information. They can effectively trade minor quality degradation for faster inference because these tasks are primarily perceptual and small texture inaccuracies are often visually negligible. In contrast, 3D generation requires the model to learn and synthesize precise geometric structures, a task where small numerical inaccuracies introduced by caching can accumulate into major inconsistencies. Unlike texture or color errors in 2D images, deviations in voxel or point-level predictions directly affect the topology and spatial integrity of the 3D object. 

For instance, in TRELLIS \cite{xiang2025structured}, a state-of-the-art 3D diffusion framework, when naive caching is applied to its geometry generation phase, even minor inaccuracies in cached voxels or latent features can produce structural artifacts such as surface holes, geometric distortions or non-manifold meshes. This highlights the need for geometry-aware caching strategies that exploit computational redundancy while maintaining integrity of both geometry and texture in 3D generation.

To address this problem, we propose \textbf{Fast3Dcache}, designed to accelerate inference while preserving structural correctness. Distinct from 2D approaches that rely on perceptual feature similarity, our method addresses the unique structural redundancy of 3D geometry. By capturing the dynamic evolution of sparse voxels, we enable aggressive computation skipping without compromising the strict topological integrity required for 3D shapes. Our approach is motivated by a key observation derived from analyzing state-of-the-art 3D diffusion frameworks TRELLIS \cite{xiang2025structured}: during the denoising process, the occupancy field, which indicates voxel existence, exhibits a progressively stabilizing pattern. Specifically, as denoising proceeds, an increasing number of voxel locations become static, meaning their occupancy values no longer change across subsequent timesteps. Moreover, the number of these active updates decreases approximately following a logarithmic pattern. Inspired by this observation, we design a dynamic caching mechanism that identifies and caches features corresponding to stable voxel regions in the latent feature space. By adaptively tuning the cache ratio based on the observed change rate, our method avoids redundant computations in static regions while focusing inference on dynamically evolving parts of the geometry. We formalize this scheduling strategy as Predictive Caching Scheduler Constraint (\textbf{PCSC}), which predicts the number of stable voxels and controls the caching ratios over timesteps effectively. 

After determining the cache quota for each timestep according to the stabilization pattern, we further design a robust selection criterion, termed the Spatiotemporal Stability Criterion (\textbf{SSC}), to accurately identify which tokens should be cached. Intuitively, given the cache quota predicted by the PCSC scheduler, our goal is to cache those tokens that have exhibited stable behavior in recent timesteps, namely, features corresponding to regions whose geometric states have largely converged.
To achieve this, SSC evaluates voxel stability from two complementary perspectives. The first is the magnitude of the predicted velocity, which reflects how much a voxel’s latent representation changes between consecutive timesteps. 
The second is the acceleration, which measures the stability of velocity through the rate of change in size and direction.
By jointly considering both magnitude and direction, SSC provides a more fine-grained measure of voxel stability than either metric alone, enabling accurate and adaptive caching decisions.

We conduct comprehensive experiments on 3D generation tasks. Our approach accelerates inference substantially and maintains high geometric quality compared to non-accelerated baselines and naive caching strategies.
Our main contributions are summarized as follows:
\begin{itemize}
 \item We propose a novel geometry-aware caching framework for 3D diffusion models, leveraging the intrinsic stabilization patterns of voxel occupancy during denoising.
 \item We design \textbf{Predictive Caching Scheduler Constraint (PCSC)} that dynamically adjusts caching ratios over timesteps based on the predicted stabilization trend.
 \item  We introduce \textbf{Spatiotemporal Stability Criterion (SSC)}, a robust token-selection rule that selects stable voxel tokens through a joint analysis of velocity magnitude and acceleration magnitude.
 \item Extensive experiments validate that our approach achieves state-of-the-art acceleration-performance trade-offs on 3D generation tasks. Code and models are publicly available.
\end{itemize}

%% file: sec/2_related_work.tex
\section{Related Work}
\label{Related Work}
\paragraph{Diffusion and Flow-based 3D Generative Models.} 
Previously, Score Distillation Sampling (SDS) \cite{poole2022dreamfusion,lin2023magic3d,chen2023fantasia3d,tang2023makeit3d,tang2024dreamgaussian,yi2023gaussiandreamer,yang2024learn} was widely utilized for 3D content creation. However, this approach suffers from significant limitations, including slow per-scene optimization speeds and multi-view inconsistencies known as the Janus problem. With the emergence of large-scale 3D datasets such as Objaverse \cite{deitke2023objaverse,lin2025objaverse++}, researchers have increasingly utilized this data to train DM or FM capable of generating 3D objects directly.

Current direct 3D generation methods utilize different underlying representations, which can be primarily categorized into explicit \cite{xiang2025structured,wu2024direct3d,wu2025direct3d}, and implicit latent sets \cite{10.1145/3658146, li2024craftsman3d,hunyuan3d22025tencent,yang2024hunyuan3d}. Compared to implicit latent vectors, explicit voxel representations offer superior control over spatial structure and topology. Among works utilizing explicit representations, TRELLIS \cite{xiang2025structured} stands out due to its distinct two-stage design, which decouples geometry synthesis from texture generation. Consequently, TRELLIS has become a foundational framework for various downstream tasks, with subsequent research addressing specific challenges: DSO \cite{li2025dso} incorporates physics-based guidance to ensure physical soundness, Amodal3R \cite{wu2025amodal3r} resolves occlusion issues, and other works focus on refinement and part-aware modeling \cite{ryu2025elevating,yang2025omnipart,zhang2025gaussian,he2025sparseflex,ye2025hi3dgen}.
\vspace{-1.5em}
\paragraph{Acceleration Works of DM / FM.} 
Acceleration for 2D or video diffusion models is broadly categorized into \emph{training-required} and \emph{training-free} approaches. (1) \emph{Training-required Methods} include distillation \cite{ma2025diffusion,kim2025autoregressive,dao2025self,salimans2022progressive,meng2023distillation} and consistency models \cite{song2023consistency,wang2025videoscene}. These methods require expensive retraining and permanently modify model weights. They are often limited by specific frameworks. (2) \emph{Training-free Methods} reduce inference costs by exploiting redundancies without altering weights. These include adaptive solvers \cite{lu2022dpm,lu2025dpm} or sampling strategies \cite{shao2025rayflow,ding2025rass,kim2025adaptive,ren2025grouping} that reduce step counts, attention optimizations \cite{shen2025draftattention,yang2025sparse,zhang2025spargeattn,bolya2022token,wu2025direct3d}, and model pruning \cite{tong2025flowcut,cai2025fastflux}. Most relevant to our work is feature caching \cite{zou2024accelerating,zhou2025less,wu2025fast,liu2025plug,selvaraju2024fora,zhang2025blockdance,ma2024deepcache,liu2025reusing,feng2025hicache,fan2025taocache,lv2024fastercache,chen2025accelerating}, which reuses features based on spatial / temporal similarity. However, these methods are predominantly designed for 2D / video tasks. Their direct migration to 3D generation often causes fatal topological errors by ignoring unique geometric characteristics. While Hash3D \cite{yang2025hash3d} explored 3D acceleration, it is not applicable to diffusion-based frameworks.

%% file: sec/3_Preliminaries.tex
\section{Preliminaries}
\label{pre}
\subsection{Flow Matching (FM)}
Many 3D generation frameworks leverage Flow Matching (FM) \cite{esser2024scaling,lipman2022flow,liu2022flow,wang2024rectified,yang2024consistency,lee2024improving,wang2025videoscene}, particularly the efficient rectified flow formulation. The ideal velocity field $\mathbf{u}_t = \mathbf{y}_1 - \mathbf{y}_0$ serves as the ground-truth, defined along the path $\mathbf{y}_t = (1-t)\mathbf{y}_0 + t\mathbf{y_1}$ that interpolates data $\mathbf{y}_0 \sim p_{\text{data}}$ and noise $\mathbf{y_1} \sim \mathcal{N}(0,\mathbf{I})$. A neural network $\mathbf{v}_\theta(\mathbf{y}_t,t)$ is trained to approximate this field. Inference generates a sample $\mathbf{y}_0$ from noise $\mathbf{y_1}$ by numerically solving the ODE $\mathbf{y}_0= \mathbf{y_1}-\int_0^1 \mathbf{v}_\theta(\mathbf{y}_t,t)\text{d}t$. \emph{The characteristics of the predicted velocity field $\mathbf{v}_\theta$ at each step $t_k$ inform our caching strategy.}

\subsection{Sparse Structure Generation}
The TRELLIS framework \cite{xiang2025structured} generates 3D assets in two stages: Structure Generation and SLAT Generation. Our work accelerates the first stage, which defines the structure as a set of active voxel coordinates $\mathbf{\mathcal{P}} = \{p_i\}_{i=1}^{L}$. A Flow Transformer $\mathbf{\mathcal{G}}_S$, conditioned on a DINOv2-processed image $c$, iteratively predicts the velocity field to evolve a noise grid $\mathbf{\mathcal{S}_\epsilon} \in \mathbb{R}^{B\times C\times D\times H \times W}$. By decoding this latent grid at each step, we observe a distinct three-phase stabilization pattern where voxel changes progressively diminish. \emph{This predictable behavior provides a clear opportunity to accelerate the Structure Generation process.}

%% file: sec/4_Method.tex
\section{Methodology}

In Sec.~\ref{observation}, we present key observations on 3D geometry synthesis that motivate our acceleration strategy. Based on these insights, Sec.~\ref{fast3Dcache Core} introduces the core components of Fast3Dcache. Finally, Sec.~\ref{fastfast} details the integration of these components into a unified, three-stage pipeline.

\subsection{3D Geometry Synthesis Observation}
\label{observation}
In TRELLIS \cite{xiang2025structured}, geometry synthesis is achieved by iteratively rectifying a latent feature grid $\mathcal{S}_t$, which is decoded at each step to determine the underlying 3D structure. 
To design a geometry-aware caching strategy, we first analyze how the generated geometry evolves over time. Our study reveals two complementary forms of redundancy: (1) three-phase stabilization pattern in voxel occupancy that follows a predictable log-linear decay, and (2) stabilization of latent features, reflected in the magnitude and temporal variation of the predicted velocity field. Together, these observations indicate that large portions of the grid become progressively stable as sampling proceeds, revealing substantial computational redundancy and suggesting that feature caching can be safely exploited in these regions.

\begin{figure}[t]
	\centering
	\begin{subfigure}{0.49\linewidth}
		\centering
		\includegraphics[width=1.0\linewidth]{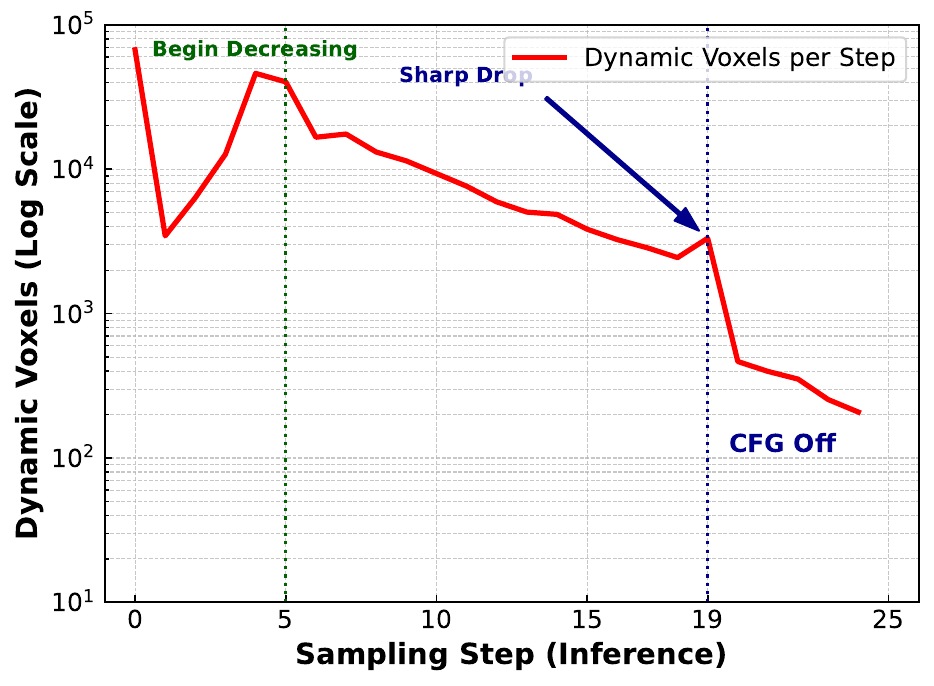}
		\caption{Original curve.}
        \label{origin}
	\end{subfigure}
    \centering
	\begin{subfigure}{0.49\linewidth}
		\centering
		\includegraphics[width=1.0\linewidth]{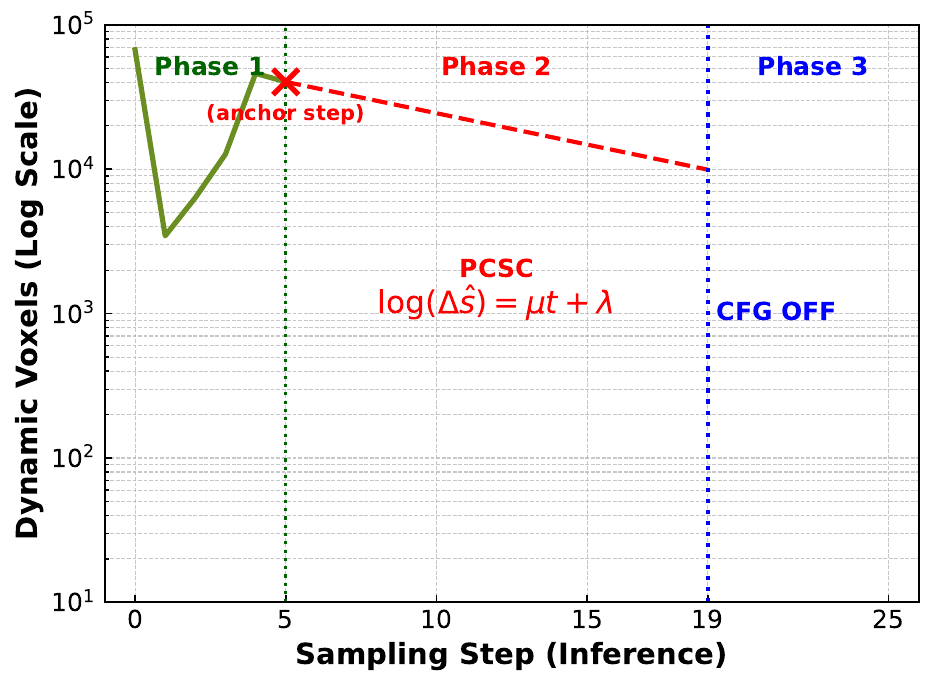}
		\caption{PCSC curve.}
        \label{pcscpcsc}
	\end{subfigure}
    \caption{\textbf{Observed voxel stabilization trend and the PCSC motivation.} (a) The \emph{Original curve} plots the empirically observed number of dynamic voxels (log-scale) per inference step, revealing a distinct three-phase pattern. (b) The \emph{PCSC curve} illustrates our approach, motivated by this observation. We identify that the decay in Phase 2 can be reliably approximated by a log-linear function (red dashed line). This predictability forms the foundation for our scheduler, which we calibrate at an anchor step to forecast the stabilization budget.}
	\label{chw}
\end{figure}

\begin{figure*}[t]
	\centering
	\begin{subfigure}{1\linewidth}
		\centering
		\includegraphics[width=1.0\linewidth]{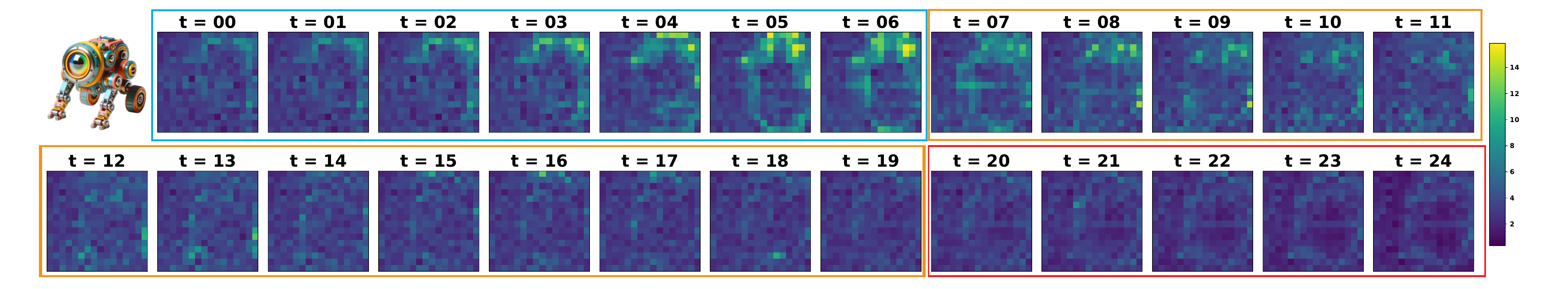}
		\caption{\textbf{Visualization of velocity field feature map.} This panel displays the temporal evolution of the predicted velocity field $\mathbf{v}_t$ for each token within a central spatial slice of the feature grid $\mathbf{\mathcal{S}}_t$.}
		\label{vvv}
	\end{subfigure}
    \qquad
    \centering
	\begin{subfigure}{1\linewidth}
		\centering
		\includegraphics[width=1.0\linewidth]{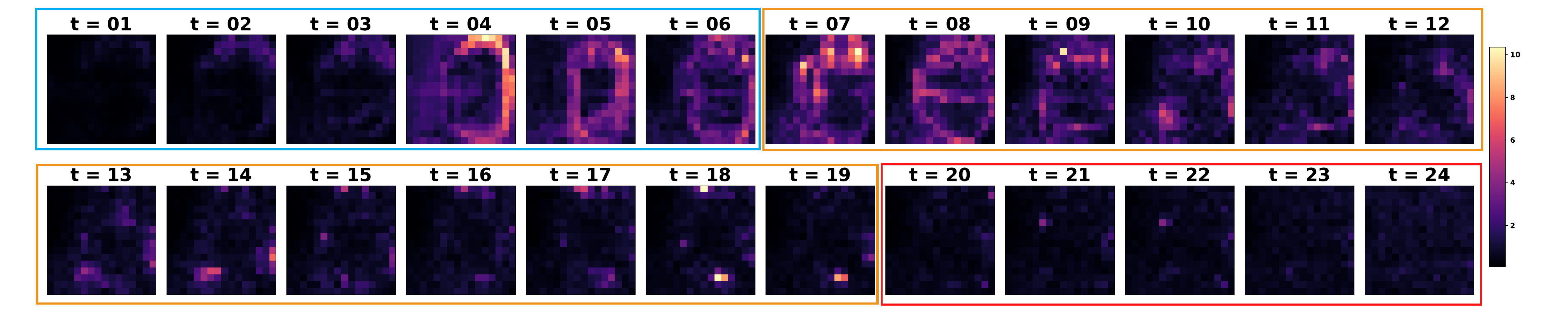}
		\caption{\textbf{Visualization of acceleration field feature map.} This panel illustrates the temporal evolution of the difference between consecutive velocity fields ($||\mathbf{v}_t - \mathbf{v}_{t-1}||_2$, acceleration), representing the instantaneous caching error for each token within the central slice.}
		\label{aaa}
	\end{subfigure}
	\caption{\textbf{Visualization of velocity field and acceleration field feature maps in $\mathbf{\mathcal{S}}_t$.} The maps illustrate the temporal dynamics of (a) velocity magnitude and (b) acceleration magnitude (rate of change). These tiny dynamics mirror the three-phase stabilization pattern observed in Fig.~\ref{origin}. The progressive decay in both velocity and acceleration magnitudes confirms their efficacy as robust criteria for identifying stable tokens suitable for caching.}
	\label{va}
\end{figure*}

\vspace{-1em}
\paragraph{Voxels Evolution in Binary 3D Grid.}
To determine when caching can be safely applied, we require a measure of geometric change across timesteps.
Instead of analyzing latent tokens directly, we decode the latent grid $\mathbf{\mathcal{S}}_t$ into a binary occupancy grid $\mathbf{\mathcal{O}}_t \in \mathbb{R}^{N^3}$. This representation allows us to quantify \emph{dynamic voxels}, \textit{i.e.} the voxels whose occupancy state changes between consecutive timesteps:

\begin{equation}\label{variable}
\Delta s_{t} = \sum_{i,j,k} \left(\mathcal{O}_{t+1}(i,j,k) \oplus \mathcal{O}_{t}(i,j,k)\right),
\end{equation}
where $i,j,k \in \{0, 1, \ldots, N-1\}$ and $\oplus$ denotes the XOR operation, which directly reflects whether each voxel has flipped its state. 
A large value of $\Delta s_t$ indicates that the global geometry is still rapidly evolving, whereas a small value suggests that the structure has largely stabilized. 

Fig.~\ref{origin} plots $\Delta s_t$ over time for TRELLIS~\cite{xiang2025structured}. We consistently observe a clear three-phase pattern: (i) an initial \emph{unstable} phase, where $\Delta s_t$ is large and the coarse geometry is being formed, (ii) an intermediate phase, where $\Delta s_t$ decreases approximately log-linearly as the geometry progressively stabilizes, (iii) a final phase, where $\Delta s_t$ drops sharply and only minor refinements occur. This phase separation is crucial for our method: it suggests that caching should be disabled in the early unstable phase, gradually introduced with a growing budget in the intermediate phase, and applied most aggressively in the final refinement phase. In other words, the voxel evolution curve provides a principled way to allocate different caching budgets to different stages of the generative process. 

\vspace{-1em}
\paragraph{Feature Dynamics in Latent Grid.} 
The voxel trend provides a global caching budget at each timestep, but it does not specify \emph{which} tokens can be safely cached. To select specific tokens, we need a more fine-grained measure of local stability in latent space. For this purpose, we examine the velocity field predicted by the network at each timestep and analyze both its magnitude and its temporal variation.

(1) \emph{Velocity Field Analysis}: We define the velocity magnitude for token $i$ at timestep $t$ as $V_i(t) = ||v_i(t)||_2$, which represents the intensity of feature updates. 
As shown in Fig.~\ref{vvv}, the distribution of $V_i(t)$ also exhibits a three-phase evolution. In Phase~1 (blue region), many tokens have large velocity magnitudes, reflecting the need for substantial updates to establish the coarse object structure. Caching in this stage would risk corrupting the emerging geometry. In Phase~2 (orange region), the number of tokens with large $V_i(t)$ gradually decreases, indicating that more regions of the grid become stable over time. In Phase~3 (red region), most tokens exhibit small velocity magnitudes, and the model only performs subtle refinements. These observations suggest that tokens with persistently small velocity magnitudes are natural candidates for caching.

(2) \emph{Acceleration Field Analysis}: We define Instantaneous Caching Error (ICE), equivalent to the acceleration magnitude $A_i(t)$, to quantify the potential error incurred by approximating the current velocity with the previous step:
\begin{equation}
\text{ICE}_i(t) \triangleq A_i(t) = ||v_i(t) - v_i(t-1)||_2. 
\end{equation}

Intuitively, $A_i(t)$ measures how much the current update direction deviates from the previous one.
In Fig.~\ref{aaa}, high-acceleration events correlate strongly with the structural changes observed in the velocity field but provide a more rigorous measure of instability. Consequently, we leverage both velocity and acceleration metrics as the joint criteria for token selection.

\begin{figure*}[t]
    \centering
    \includegraphics[width=1.0\linewidth]{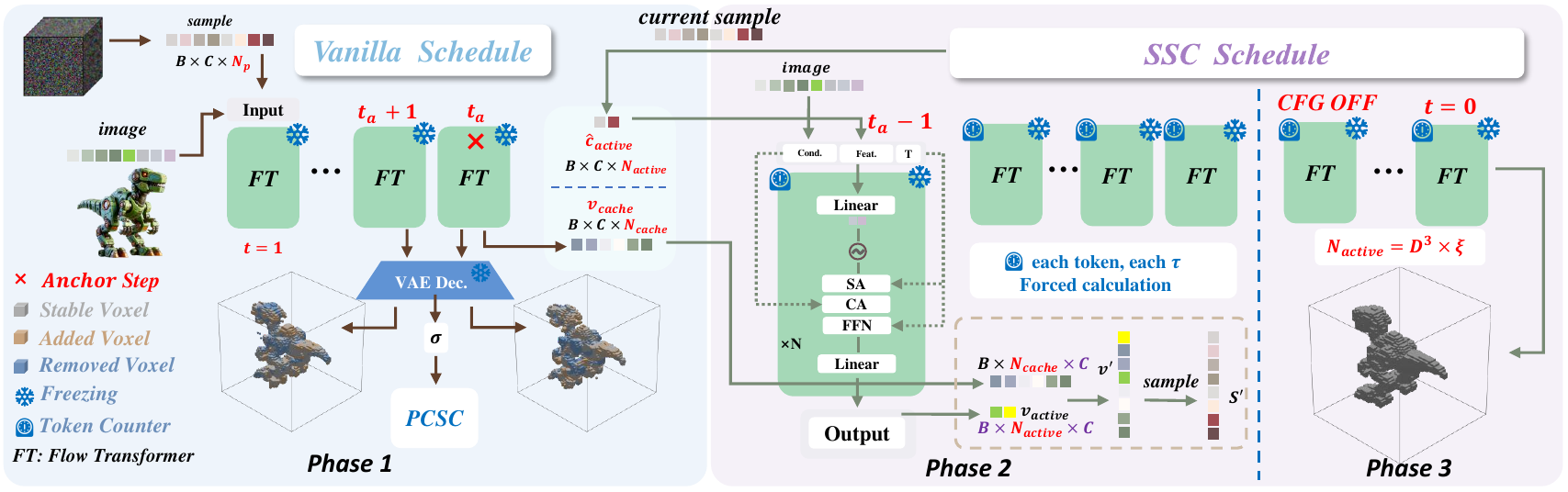}
\caption{\textbf{Overview of the Fast3Dcache three-stage acceleration strategy.} \textbf{Phase 1 (Full Sampling):} The process begins with full sampling to establish initial geometric stability. At the end of this phase, the \textbf{PCSC} is calibrated by measuring voxel change ($\sigma$) at the anchor step. \textbf{Phase 2 (Dynamic Caching):} In the main phase, the \textbf{SSC} identifies stable tokens for caching based on the dynamic budget predicted by PCSC. Only unstable tokens are processed by the FT. \textbf{Phase 3 (CFG-Free Refinement):} The final stage employs an aggressive fixed-ratio schedule. A high and fixed ratio $\xi$ is used to determine the proportion of tokens to cache, maximizing computational savings during these stable refinement steps.}
    \label{pipeline}
\end{figure*}

\subsection{Fast3Dcache Core Components}
\label{fast3Dcache Core}
Building upon the observations, we introduce two complementary components that jointly determine the caching strategy in \emph{Fast3Dcache}. The \emph{Predictive Caching Scheduler Constraint (PCSC)} specifies \emph{how many} tokens may be cached at each timestep, while the \emph{Spatiotemporal Stability Criterion (SSC)} determines \emph{which} specific tokens can be safely cached without degrading geometric fidelity. Together, these modules translate geometric evolution into a dynamic and fine-grained computational policy.
\vspace{-1em}
\paragraph{Predictive Caching Scheduler Constraint (PCSC).} 
The goal of PCSC is to allocate an appropriate caching budget at each timestep. 
Motivated by the distinct stabilization pattern observed in Phase 2 of Fig.~\ref{origin}, we approximate the decline in dynamic voxels using a \emph{log-linear} curve. This predictive approach enables the model to dynamically determine the optimal cache quota at each timestep, ensuring that the computational budget adapts flexibly to the evolving stability of the geometry.

To construct the decay schedule efficiently with minimal computational overhead, we perform a one-time calibration at the end of Phase 1. We designate a specific timestamp as the \emph{anchor step}, calculated as $ \lceil T\times \rho_\text{a} \rceil$, where $T$ is the total number of inference steps and $\rho_\text{a}$ governs the duration of the full-sampling stage. At this anchor, we quantify the initial magnitude of voxel changes, denoted as $\sigma$, by comparing the decoded grids of adjacent steps. Empirically, the rate at which dynamic voxels decay is consistent across diverse samples, allowing us to adopt a fixed slope parameter $\mu$ to extrapolate future changes. As illustrated in Fig.~\ref{pcscpcsc}, we model the decay of dynamic voxels as a straight line in log-coordinate system: 
\begin{equation}
\log(\Delta \hat{s}) = \mu \cdot t + \lambda,
\end{equation}
where $\Delta \hat{s}$ is the predicted dynamic voxel between adjacent timesteps and $\lambda$ is the vertical intercept. {The line in log-coordinate yields final predictive curve:}
\begin{equation}
\Delta \hat{s}= \sigma \cdot e^{\mu\cdot (t - \lceil T\cdot \rho_\text{a} \rceil)}.
\end{equation}

This predictive curve provides a time-varying estimate of geometric change $\Delta \hat{s}_t$ across intermediate timesteps. To translate this geometric prediction into a computational budget for the flow transformer, we derive the number of tokens to be cached $c_t$. Since the dynamic voxels are defined in the upsampled output space, we normalize them by the upsampling factor $\gamma_\text{up}$ to estimate the necessary active calculations, and designate the remainder as the cache quota: 
\begin{equation}
c_t = D^3 - \frac{\Delta \hat{s}_t}{\gamma_\text{up}},
\end{equation}
where $D^3$ represents the total number of latent tokens and $\gamma_\text{up}$ denotes the upsampling ratio. This derived $c_t$ serves as the dynamic caching budget that strictly constrains the token selection process in the subsequent Phase 2.
This dynamic budget specifies an upper bound on the number of cached tokens for Phase~2 and ensures that caching aggressiveness adapts to the stability of the underlying geometry.

\vspace{-1em}
\paragraph{Spatiotemporal Stability Criterion (SSC).}
While PCSC establishes the global cache budget $c_t$, the complementary challenge is to pinpoint exactly which tokens can be safely cached without compromising geometric fidelity. This selection process requires a metric that is both accurate and computationally lightweight. 
To achieve this, we introduce the \emph{Spatiotemporal Stability Criterion (SSC)}, which evaluates token-wise stability based on instantaneous velocity and acceleration dynamics. SSC is applied throughout Phases~2 and~3 to distinguish tokens that require fresh computation from those whose features have converged.

Guided by the observations in Sec.~\ref{observation}, we define for each token $i$ a \emph{cache ability score} $C_i(t)$ that integrates two normalized quantities: the acceleration $A_i(t)$ (representing temporal variation of updates) and the velocity magnitude $V_i(t)$ (representing update intensity). Formally,
\begin{equation}
\label{ssc}
    C_i(t) = \omega \cdot \mathrm{norm}\!\left(A_i(t)\right)
           + (1 - \omega) \cdot \mathrm{norm}\!\left(V_i(t)\right),
\end{equation}
where
\vspace{-1em}
\begin{equation}\label{eq:normalize}
  \begin{aligned}
    \mathrm{norm}\!\bigl(A_i(t)\bigr)
      &= \frac{A_i(t) - \min_j A_j(t)}{\max_j A_j(t) - \min_j A_j(t)}, \\
    \mathrm{norm}\!\bigl(V_i(t)\bigr)
      &= \frac{V_i(t) - \min_j V_j(t)}{\max_j V_j(t) - \min_j V_j(t)}.
  \end{aligned}
\end{equation}

Intuitively, this score captures how unstable a token is: tokens with small, slowly-varying updates are more stable and therefore more suitable for caching, whereas tokens with large or rapidly-changing updates are less stable and should be recomputed more frequently.

The 3D latent state $\mathbf{\mathcal{S}}_t$ is first flattened into a dense sequence of tokens $\mathbf{x}^{(t)} \in \mathbb{R}^{B \times N_p \times d_\text{model}}$, where $N_p = D \times H \times W$ denotes the total number of tokens. Given a computational budget $c_t$ (the maximum number of tokens we can afford to actively update at step $t$), the SSC ranks tokens according to their cache ability scores and identifies the indices of the active subset, denoted by $\mathcal{I}^{(t)}_\text{active}$. We apply an index-based selection to extract the corresponding features and obtain a reduced input sequence $\mathbf{x}^{(t)}_\text{active} = \mathbf{x}^{(t)}[:, \mathcal{I}^{(t)}_\text{active},:]$, which contains only the tokens chosen under the budget $c_t$. Since the attention mechanism~\cite{vaswani2017attention} is the primary computational bottleneck in the generation process, we perform self-attention exclusively on this active subset. Only the less stable, more informative tokens are recomputed, while the more stable tokens reuse their cached states. This stability-aware selection allows us to respect the budget $c_t$ and substantially reduce the cost of attention.

\subsection{Fast3Dcache Integration} \label{fastfast} Having established the PCSC budget scheduler and the SSC token selector in Sec.~\ref{fast3Dcache Core}, we now integrate these components into a unified, end-to-end acceleration workflow Fast3Dcache. As illustrated in Fig.~\ref{pipeline}, Fast3Dcache segments the inference process into three strategic phases. This multi-stage design balances the necessity for geometric stability in the early steps with the opportunity for aggressive acceleration in the later convergent stages.

\begin{itemize} \item \textbf{Phase 1: Full Sampling.} Consistent with observations in other generative modalities \cite{jeong2025upsample,tian2025training,zhang2025blockdance,liu2025region}, the initial steps of 3D generation exhibit high volatility in voxel evolution. Consequently, we employ full sampling during this phase to guarantee fundamental geometric accuracy. Crucially, the final step of this phase serves as the \emph{anchor point} to calibrate the PCSC scheduler, allowing us to predict the decay trajectory and determine the cache budget for the subsequent phase.

\item \textbf{Phase 2: Dynamic Caching.} In this intermediate phase, we deploy the SSC module to execute precise, token-level caching based on the dynamic budget provided by PCSC. However, relying exclusively on cached features for extended periods can lead to geometric errors. To mitigate this, we enforce an \emph{Error Accumulation Elimination} constraint defined by the interval $\tau$, where $\tau$ is the interval controlling the frequency of full refresh steps. 
We mandate a full-sampling update every $\tau$ steps to rectify the latent states and limit the propagation of approximation errors.
This periodic reset keeps the latent grid aligned with the correct generative trajectory.

\item \textbf{Phase 3: CFG-Free Refinement.} As shown in Fig.~\ref{origin}, the generation process enters a highly stable regime once Classifier-Free Guidance (CFG) \cite{ho2022classifier} is disabled, focusing primarily on minor structural refinements. To streamline efficiency during this stage, we transition to a simplified fixed-ratio caching strategy. We continue to utilize the SSC for token selection, but the cache budget $c_t$ is governed by a constant, aggressive ratio $\xi$. We can calculate it as $c_t = D^3 \cdot \xi$, where $D^3$ denotes the total token volume. To counteract potential error accumulation over this extended sequence, we introduce a periodic correction cycle governed by the parameter $f_\text{corr}$. The model operates in a cached mode for $f_\text{corr}-1$ steps, followed by a \emph{Full Correction Step} every $f_\text{corr}$-th step, where all tokens are recalculated to fully realign the feature grid.
\end{itemize}

\definecolor{highlightgray}{gray}{0.9}
\begin{table*}[t]
\centering
\caption{\textbf{Quantitative comparison on TRELLIS \cite{xiang2025structured} and DSO \cite{li2025dso} frameworks.} We benchmark Fast3Dcache against TRELLIS and existing modality-aware method (RAS \cite{liu2025region}). Our method consistently outperforms the baseline, achieving higher throughput and lower FLOPs while preserving geometric fidelity (CD and F-Score) across various settings. (\textbf{best} and \underline{second-best})}
\label{table1}
\resizebox{\textwidth}{!}{%
\begin{tabular}{llcccc}
\toprule
\textbf{Frameworks} & \textbf{Acceleration Methods} & \textbf{Throughput (iter/s)$\uparrow$}  & \textbf{FLOPs (T)$\downarrow$} & \textbf{CD$\downarrow$} & \textbf{F-Score$\uparrow$} \\
\midrule
TRELLIS \cite{xiang2025structured} & vanilla & 0.5055 & 244.2 & 0.0686 & 54.8244 \\
\midrule
& RAS \cite{liu2025region} (sample ratio 25\%) & 0.6337 (\textcolor{orange}{$\uparrow 25.36$ \%})& 125.1 (\textcolor{blue}{$\downarrow 48.77$ \%})& 0.0867 (\textcolor{red}{$\uparrow 26.38$ \%})& 40.2769 (\textcolor{red}{$\downarrow 26.53$ \%})\\

& RAS \cite{liu2025region} (sample ratio 12.5\%) & 0.6177 (\textcolor{orange}{$\uparrow 22.20$ \%})& 125.8 (\textcolor{blue}{$\downarrow 48.48$ \%})& 0.0846 (\textcolor{red}{$\uparrow 23.32$ \%})& 43.9622 (\textcolor{red}{$\downarrow 19.81$ \%})\\

\rowcolor{highlightgray} & \textbf{Fast3Dcache ($\tau=3$)} & 0.5850 (\textcolor{orange}{$\uparrow 15.73$ \%})& 142.4 (\textcolor{blue}{$\downarrow 41.69$ \%})& \textbf{0.0697} (\textcolor{red}{$\uparrow 1.60$ \%}) & \textbf{54.0900} (\textcolor{red}{$\downarrow 1.34$ \%}) \\

\rowcolor{highlightgray} & \textbf{Fast3Dcache ($\tau=5$)} & \underline{0.6344} (\textcolor{orange}{$\uparrow 25.50$ \%})& \underline{121.3} (\textcolor{blue}{$\downarrow 50.33$ \%}) & 0.0712 (\textcolor{red}{$\uparrow 3.79$ \%}) & 53.5003 (\textcolor{red}{$\downarrow 2.42$ \%}) \\

\rowcolor{highlightgray} & \textbf{Fast3Dcache ($\tau=8$)} & \textbf{0.6426} (\textcolor{orange}{$\uparrow 27.12$ \%})& \textbf{110.3} (\textcolor{blue}{$\downarrow 54.83$ \%}) & \underline{0.0703} (\textcolor{red}{$\uparrow 2.48$ \%}) & \underline{53.7528} (\textcolor{red}{$\downarrow 1.95$ \%})\\

\midrule
DSO \cite{li2025dso}  & vanilla & 0.3496 & 244.2 & 0.0687 & 54.8350 \\
\midrule
& RAS \cite{liu2025region} (sample ratio 25\%) &\textbf{0.4341} (\textcolor{orange}{$\uparrow 24.17$ \%}) & \underline{125.0} (\textcolor{blue}{$\downarrow 48.81$ \%}) & 0.0805 (\textcolor{red}{$\uparrow 17.18$ \%}) & 46.4990 (\textcolor{red}{$\downarrow 15.20$ \%})\\

& RAS \cite{liu2025region} (sample ratio 12.5\%) & 0.4047 (\textcolor{orange}{$\uparrow 15.76$ \%})& 125.8 (\textcolor{blue}{$\downarrow 48.48$ \%})& 0.0820 (\textcolor{red}{$\uparrow 19.36$ \%}) & 45.5584 (\textcolor{red}{$\downarrow 16.92$ \%})\\

\rowcolor{highlightgray} & \textbf{Fast3Dcache ($\tau=3$)} & 0.3955 (\textcolor{orange}{$\uparrow 13.13$ \%})& 146.5 (\textcolor{blue}{$\downarrow 40.01$ \%}) & \textbf{0.0698} (\textcolor{red}{$\uparrow 1.60$ \%}) & \textbf{54.0451} (\textcolor{red}{$\downarrow 1.44$ \%})\\

\rowcolor{highlightgray} & \textbf{Fast3Dcache ($\tau=5$)} & \underline{0.4114}  (\textcolor{orange}{$\uparrow 17.68$ \%})& 126.0 (\textcolor{blue}{$\downarrow 48.40$ \%})& 0.0711 (\textcolor{red}{$\uparrow 3.49$ \%}) & \underline{53.5506} (\textcolor{red}{$\downarrow 2.34$ \%})\\

\rowcolor{highlightgray} & \textbf{Fast3Dcache ($\tau=8$)} & 0.4071 (\textcolor{orange}{$\uparrow 16.45$ \%}) & \textbf{115.4} (\textcolor{blue}{$\downarrow 52.74$ \%})&\underline{0.0704} (\textcolor{red}{$\uparrow 2.47$ \%}) & 53.5487 (\textcolor{red}{$\downarrow 2.35$ \%})\\

\bottomrule
\end{tabular}
}
\end{table*}

%% file: sec/5_Experiment.tex
\section{Experiments}\label{experiment}
In this section, we present experimental details and evaluations to demonstrate the effectiveness of our method. Sec.~\ref{imdetail} introduces the implementation details, while Sec.~\ref{results} provides quantitative results, qualitative visualizations, and ablation studies.

\subsection{Implementation Details}
\label{imdetail}
\paragraph{Setting.}
Our experiments are conducted on TRELLIS \cite{xiang2025structured} and its variant DSO \cite{li2025dso}, focusing on accelerating the inference in the initial sparse structure generation stage. A single NVIDIA GeForce RTX 4090 GPU is used in our experiment. To ensure fairness across all methods, we use FlashAttention by default in all our experiments.

\vspace{-1em}
\paragraph{Evaluation.}
To evaluate acceleration in geometry generation, we measure throughput (iters/s) and FLOPs (T) for inference efficiency. For geometric fidelity, we adopt Chamfer Distance (CD) and F-Score (threshold = 0.05), following standard protocols in 3D generation \cite{liu2023one,melas2023realfusion,wang2024crm}. All generated meshes are normalized into a unit cube and aligned with ground truth using the Iterated Closest Point (ICP) algorithm prior to metric computation. We evaluate on the Toys4K dataset \cite{stojanov2021using} following TRELLIS~\cite{xiang2025structured}. For each object, we select its corresponding mesh as ground truth, render it from 12 fixed viewpoints, apply background removal, and filter out low-quality images. This yields 71 objects with 852 valid image prompts. Each image is fed independently into the model, and we report mean metrics across all samples for fair comparison.

\subsection{Results Analysis}
\label{results}
\paragraph{Quantitative Results.} 
Table~\ref{table1} reports a comprehensive comparison between \emph{Fast3Dcache} and several methods. In addition to the vanilla TRELLIS and DSO configurations, we include a modality-aware caching method, RAS~\cite{liu2025region}, originally designed for 2D DiT models. Across all metrics, Fast3Dcache delivers substantial efficiency gains while maintaining high geometric fidelity. By contrast, the 3D-adapted RAS method fails to preserve structural integrity and leads to significant artifacts, a 26.53\% drop in F-Score on TRELLIS. This performance gap underscores a key observation: caching strategies developed for 2D image synthesis do not directly generalize to 3D geometry, as they overlook the distinct stabilization patterns and topology-sensitive dynamics of volumetric structures. By explicitly modeling these 3D-specific behaviors through PCSC and SSC, Fast3Dcache achieves better acceleration-quality trade-off. The results validate that geometry-aware caching is essential for reliable and efficient 3D generative modeling, and that the proposed method provides both principled and practical advantages over existing 2D techniques.

\vspace{-1em}
\paragraph{Complementarity with Modality-agnostic Accelerators.}
We further examine whether \emph{Fast3Dcache} can serve as a complementary module to existing state-of-the-art, modality-agnostic acceleration methods. To this end, we integrate Fast3Dcache with TeaCache~\cite{liu2025timestep}, a leading training-free accelerator originally developed for video diffusion. Since TeaCache is not tailored for 3D geometry, we first adapt its timestep-based caching mechanism to the 3D sparse transformer architecture.
As shown in Table~\ref{tablecom}, the 3D-adapted TeaCache alone achieves a  2.84$\times$ speedup. When combined with our geometry-aware Fast3Dcache, the acceleration further improves to 3.41$\times$ throughput, demonstrating that the two methods provide complementary gains. Remarkably, the combined approach also yields improved geometric fidelity, achieving better CD and F-Score scores than TeaCache~\cite{liu2025timestep} alone, indicating that Fast3Dcache contributes not only additional efficiency but also stabilizes geometric updates during sampling.
These results confirm that Fast3Dcache is highly compatible with existing accelerators and can be seamlessly integrated to produce compounding improvements in both speed and quality.
\vspace{-1em}

\begin{table}[t]
\centering
\caption{\textbf{Results of Fast3Dcache combined with a modality-agnostic SOTA method.} Integrating our method with the modality-agnostic acceleration framework TeaCache yields further speedup while also improving reconstruction quality.}
\label{tablecom}
\resizebox{\columnwidth}{!}{
\begin{tabular}{lccc}
\toprule
\textbf{Acceleration Methods} & \textbf{Throughput (iters/s)} $\uparrow$ & \textbf{CD$\downarrow$} & \textbf{F-Score$\uparrow$ }\\
\midrule
Vanilla & 0.51 (\textcolor{orange}{1.00$\times$})& 0.0686 & 54.8244 \\
TeaCache \cite{liu2025timestep} & 1.45 (\textcolor{orange}{2.84$\times$}) & 0.0705 & 53.5567 \\
\rowcolor{highlightgray} \textbf{TeaCache + ours} & \textbf{1.74 (\textcolor{orange}{3.41$\times$})} & \textbf{0.0701} & \textbf{53.9420} \\
\bottomrule
\end{tabular}
} 
\end{table}

\begin{figure*}[t]
    \centering
    \includegraphics[width=0.8\linewidth]{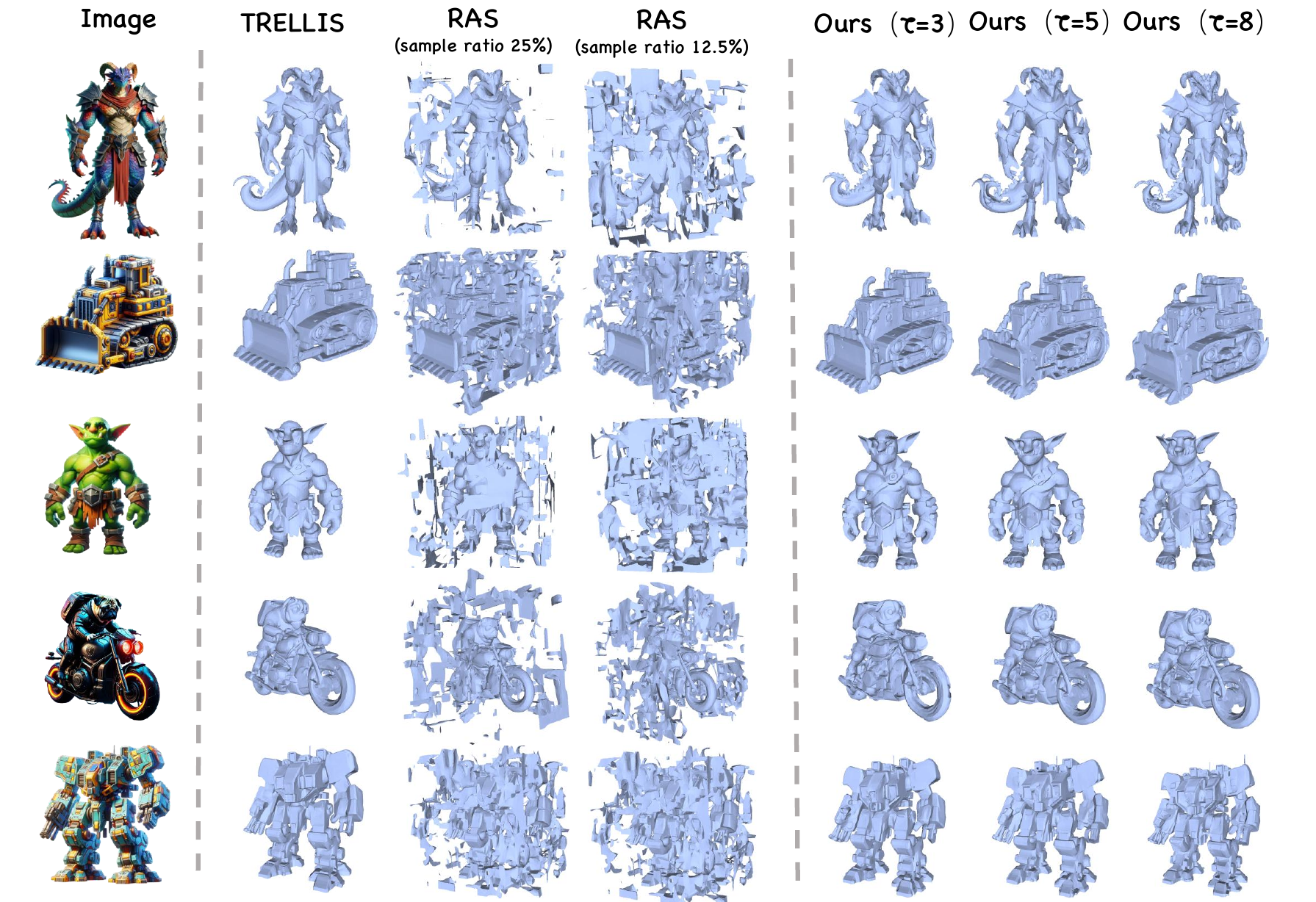}
    \caption{\textbf{Visualization comparison of 3D geometry synthesis.} The leftmost column presents the input image. Subsequent columns display 3D meshes generated by TRELLIS, RAS method (at varying sampling ratios). Observe that while RAS introduces noticeable geometric artifacts and surface noise, Fast3Dcache preserves structural fidelity comparable to the original TRELLIS framework, achieving acceleration without compromising quality.}
    \label{vi}
\end{figure*}

\paragraph{Visualization Results.} We present some 3D generation results in Fig.~\ref{vi}. Our results demonstrate that Fast3Dcache preserves structural fidelity better than the existing approach RAS \cite{liu2025region}. 

\vspace{-1em}
\paragraph{Ablation Study.} \textbf{Impact of PCSC Scheduler.} Table~\ref{pcscab} presents ablation analysis for the PCSC module. To validate the necessity of our adaptive approach, we compare PCSC against static, non-adaptive strategies with fixed sampling rates. The results demonstrate that PCSC significantly outperforms fixed-rate methods by dynamically tailoring the cache budget to the specific geometric complexity of each input prompt. Furthermore, we investigate the sensitivity of the slope parameter $\mu$. Since the cache quota $c_t$ is discretized, minor fluctuations (\textit{e.g.}, $\pm 10\%$) have a negligible effect on the final budget. Consequently, we vary the slope by an order of magnitude (factor of 10) to clearly delineate the impact of the decay rate on generation quality.

\begin{table}[t]
  \centering
  \small
    \caption{\textbf{Ablation study of the PCSC module.} We evaluate the effectiveness of our adaptive scheduler compared to fixed-rate sampling methods. Additionally, we analyze the sensitivity of the decay slope $\mu$, demonstrating that optimal slope calibration is essential for preserving generation quality.}
  \label{pcscab}
  \begin{tabular}{lccc}
    \toprule
    \textbf{Schedule} & \textbf{CD$\downarrow$} & \textbf{F-Score$\uparrow$} \\ 
    \midrule
    w.o. PCSC (fixed 25\%) & 0.0956 & 34.5122 \\
    w.o. PCSC (fixed 12.5\%) & 0.0899 & 39.0563 \\
    \midrule
    PCSC ( $\mu = -0.7$ ) & 0.0707 & 53.4978 \\
    \rowcolor{highlightgray} PCSC ( $\mu = -0.07$ ) & \textbf{0.0697} & \textbf{54.0900} \\
    PCSC ( $\mu = -0.007$ ) & 0.0701 & 53.7803\\
    \bottomrule
  \end{tabular}
\end{table}
\vspace{-1em}
\paragraph{Effectiveness of SSC Components.} Table~\ref{SSCab} details ablation analysis of SSC module. In Eq.~\ref{ssc}, the caching score is a weighted fusion of velocity ($V_i$) and acceleration ($A_i$). We evaluate the contribution of each component against the RAS method (Row 1), which utilizes standard deviation for screening. We further test single-component settings where only $V_i$ or $A_i$ is active. The results confirm that neither component alone is sufficient. The best performance is achieved by jointly considering both metrics, validating their complementary role in assessing geometric stability.

\begin{table}[t] 
  \small
  \centering
\caption{\textbf{Ablation study of the SSC module.} We evaluate the individual contributions of the velocity ($V_i$) and acceleration ($A_i$) components. The results demonstrate that relying on a single metric is insufficient, while the joint consideration of both fields yields better geometric fidelity.}
  \label{SSCab}
  \begin{tabular}{ccccc}
    \toprule
    w. $V_i(t)$ & w. $A_i(t)$ & \textbf{$\omega$} & \textbf{CD$\downarrow$} & \textbf{F-Score$\uparrow$} \\
    \midrule
          \ding{55}   & \ding{55}   & --       &      0.0743  & 50.9974   \\
          \ding{51}   & \ding{55}   & --       & 0.0836  & 44.9630   \\
          \ding{55}   & \ding{51}   & --       & 0.0709  & 53.5394   \\
    \midrule
          \ding{51}   & \ding{51}   & 0.3      & 0.0703  & 53.9156   \\
          \ding{51}   & \ding{51}   & 0.4      & 0.0706  & 53.5711   \\
          \ding{51}   & \ding{51}   & 0.5      & 0.0703  & 53.7132   \\
          \ding{51}   & \ding{51}   & 0.6      & 0.0705  & 53.8326   \\
     \rowcolor{highlightgray}  \ding{51}   & \ding{51}   & \textbf{0.7}      & \textbf{0.0697}  & \textbf{54.0900}   \\
    \bottomrule
  \end{tabular}
\end{table}
\vspace{-1em}
\paragraph{Effectiveness of Elimination Step.} Finally, we validate the critical role of the elimination step $\tau$ (Table~\ref{table1}). The results demonstrate that this constraint is indispensable for maintaining stability. Completely disabling the correction mechanism leads to significant geometric degradation, with CD deteriorating to 0.0724 and F-Score dropping to 51.8157. This confirms that periodic full-sampling updates are essential to rectify accumulated approximation errors and preserve high-fidelity generation. These experiments all demonstrate the effectiveness of each of our modules.

%% file: sec/6_Conclusion.tex
\section{Conclusion}
\label{cl}
We present Fast3Dcache, a training-free acceleration framework tailored for the TRELLIS series to expedite 3D geometry synthesis. Fast3Dcache is explicitly designed for spatially explicit representations (sparse voxels) by leveraging their inherent spatial redundancy and stabilization patterns. Our approach exploits intrinsic stabilization patterns within the generation process through two synergistic modules: Predictive Caching Scheduler Constraint (PCSC), which dynamically allocates the computational budget based on voxel decay trends, and Spatiotemporal Stability Criterion (SSC), which precisely identifies the minimal subset of active tokens requiring updates. Extensive experiments demonstrate that Fast3Dcache significantly reduces the FLOPs while strictly preserving geometric fidelity, offering a robust solution for high-quality 3D generation.

\section*{Acknowledgement}
This work was supported by the National Natural Science Foundation of China (No. 6250070674) and the Zhejiang Leading Innovative and Entrepreneur Team Introduction Program (2024R01007).

%% file: sec/X_suppl.tex
\clearpage
\setcounter{page}{1}
\maketitlesupplementary

\vspace{1cm}
\section{More Results}\label{more experiment}

\subsection{More Experiment Details}
\begin{itemize}
    \item About RAS, we re-implement RAS by directly extending it to 3D grids, following original designs (\textit{e.g.}, standard deviation and starvation prevention). We use the default sampling ratios (25\% and 12.5\%) as they are confirmed to be optimal in our experiments. 
    \item About framework DSO \cite{li2025dso}, Fast3Dcache is applied in the same way as with TRELLIS without modifications.
\end{itemize}

\subsection{Full Results of Complementarity with Modality-Agnostic Accelerators} 
To evaluate the extensibility of our approach, we integrated Fast3Dcache with existing SOTA acceleration methods, specifically TeaCache \cite{liu2025timestep} and EasyCache \cite{zhou2025less}. As presented in Table~\ref{whole_table}, the combination yields substantial performance gains. For TeaCache, integrating our method boosts the speedup to 3.41$\times$ while simultaneously surpassing the geometric quality of the standalone baseline. The results are even more pronounced with EasyCache, where the combined framework achieves a remarkable 10.33$\times$ acceleration while maintaining a high F-Score (54.77). These findings confirm that Fast3Dcache is orthogonal to existing caching strategies, enabling compounding efficiency improvements without compromising generation quality.

\begin{table}[t]
  \centering
  \caption{\textbf{Full quantitative results of Fast3Dcache combined with a modality-agnostic method.} Our combined methods obtain superior results in terms of speed and geometry quality.}
  \label{whole_table}
  \resizebox{\columnwidth}{!}{
  \begin{tabular}{llll}
    \toprule
    \textbf{Method} & \textbf{Throughput$\uparrow$ (Iters/s)} & \textbf{CD$\downarrow$} & \textbf{F-Score$\uparrow$} \\
    \midrule
    Vanilla & 0.51 & 0.0686 & 54.8244 \\
    \midrule
    TeaCache & 1.45 (2.84$\times$) & 0.0705 & 53.5567 \\
    \textbf{TeaCache + ours} & \textbf{1.74 (3.41$\times$)} & \textbf{0.0701} & \textbf{53.9420} \\
    \midrule
    EasyCache & 1.95 (3.82$\times$) & \textbf{0.0692} & 54.5051 \\
    \textbf{EasyCache + ours} & \textbf{5.27 (10.33$\times$)} & 0.0694 & \textbf{54.7722} \\
    \bottomrule
  \end{tabular}}
\end{table}

\subsection{More Results of Other Dataset and Task}
Table \ref{add_exper} shows the results of text-to-3D task on framework TRELLIS-text-xlarge and Table \ref{omni} demonstrates the results of dataset Omniobject3D \cite{wu2023omniobject3d} (216 (objects) × 4 (views) images) . For each object in Omniobject3D \cite{wu2023omniobject3d}, we select its corresponding mesh as ground truth and render it from 4 fixed viewpoints. This yields 216 objects with 864 images. While RAS causes severe quality drops (\textit{e.g.}, F-Score dropping to 47.0901 and CLIP to 13.0199) , our method ($\tau=8$) achieves significant acceleration—reaching 0.4781 and 0.6775 Iter/s across tasks—with negligible quality loss. It maintains a high CLIP score (19.3325) and preserves geometry (F-Score 73.9062, closely matching Vanilla's 74.7461).

\definecolor{highlightgray}{gray}{0.9}
\begin{table}[t]
\centering
\caption{\textbf{Results of text-to-3D on TRELLIS-text-xlarge.}}
\label{add_exper}
\resizebox{\columnwidth}{!}{%
\begin{tabular}{lccc}
\toprule
\textbf{Methods} & \textbf{Speed (Iter/s)}$\uparrow$ & \textbf{FLOPs}$\downarrow$ & \textbf{CLIP}$\uparrow$ \\
\midrule
TRELLIS Vanilla & 0.3507 & 374.5 & 22.9684  \\
\midrule
RAS (sample 25\%) & \textbf{0.5235} & 169.3 & 13.0199 \\
RAS (sample 12.5\%) & \underline{0.5224} & \textbf{143.1} & 12.9782 \\
\rowcolor{highlightgray} 
\textbf{Ours ($\tau=3$)} & 0.4622  &  204.7 & \textbf{20.2812}\\
\rowcolor{highlightgray} 
\textbf{Ours ($\tau=5$)} & 0.4751  & 171.5 & \underline{19.7698} \\
\rowcolor{highlightgray} 
\textbf{Ours ($\tau=8$)} & 0.4781 & \underline{154.4}  & 19.3325 \\
\bottomrule
\end{tabular}
}
\end{table}

\definecolor{highlightgray}{gray}{0.9}
\begin{table}[t]
\centering
\caption{\textbf{Results of dataset OmniObject3D \cite{wu2023omniobject3d} on image-to-3D.}}
\label{omni} 
\resizebox{\columnwidth}{!}{%
\begin{tabular}{lcccc}
\toprule
\textbf{Methods} & \textbf{Speed (Iter/s)}$\uparrow$ & \textbf{FLOPs}$\downarrow$ & \textbf{CD}$\downarrow$ & \textbf{F-Score}$\uparrow$ \\
\midrule
TRELLIS Vanilla & 0.5253 & 244.2 & 0.0425 & 74.7461 \\
\midrule
RAS (sample 25\%) & \underline{0.6653}  & 132.8 & 0.0761 & 47.0901 \\

RAS (sample 12.5\%) & 0.6545 & 123.9 & 0.0797 & 43.9043 \\

\rowcolor{highlightgray} 
\textbf{Ours ($\tau=3$)} & 0.6294 & 133.3  & \underline{0.0428} & \underline{74.2281} \\

\rowcolor{highlightgray} 
\textbf{Ours ($\tau=5$)} & 0.6600 & \underline{121.7}  & \textbf{0.0424} & \textbf{74.5931} \\

\rowcolor{highlightgray}  
\textbf{Ours ($\tau=8$)} & \textbf{0.6775} & \textbf{110.7} & 0.0433 & 73.9062 \\

\bottomrule
\end{tabular}
}
\end{table}

\begin{figure}[t]
    \centering
    \includegraphics[width=0.9\linewidth]{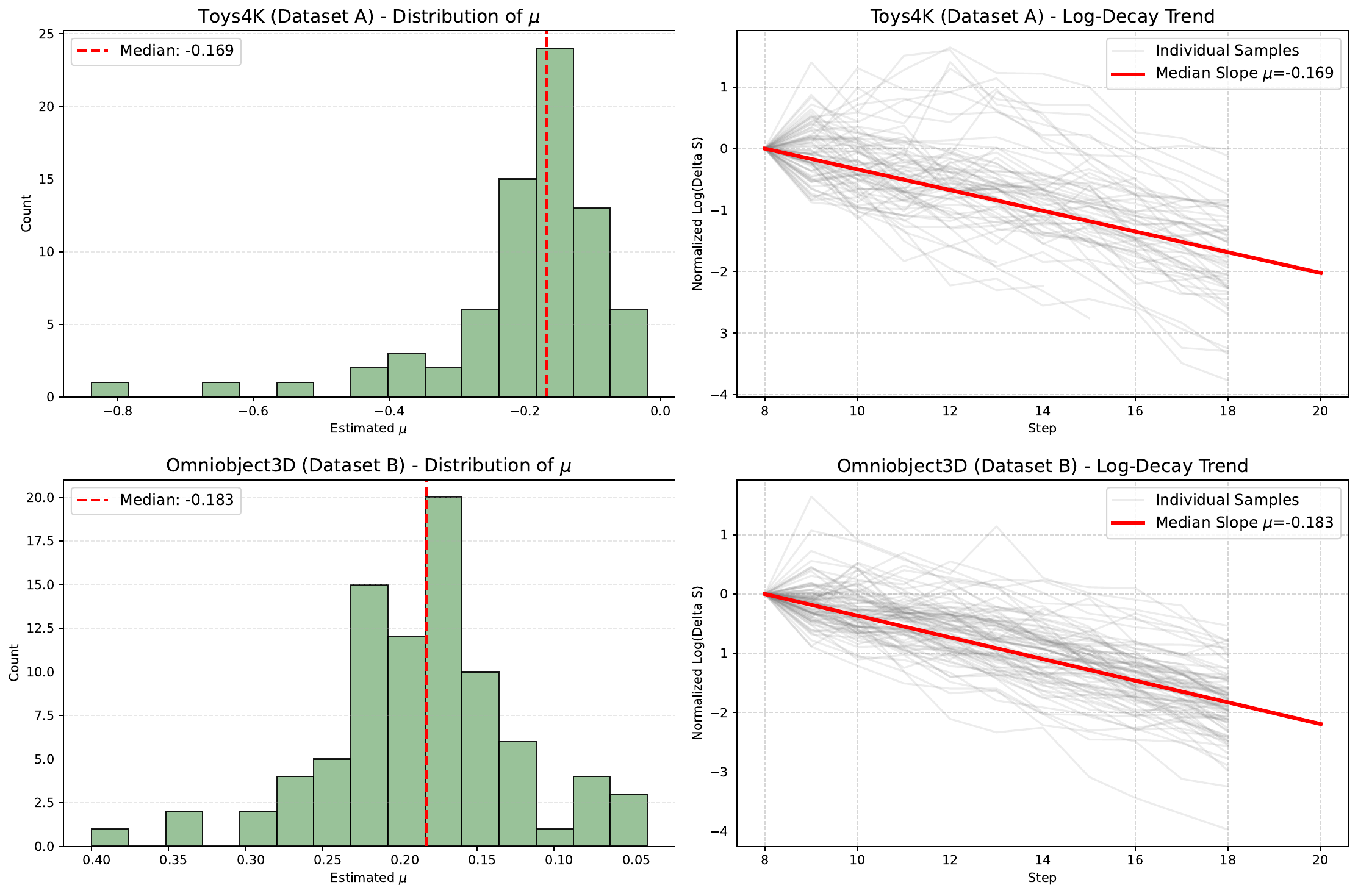}
    \caption{\textbf{Data-driven parameter selection.} Distributions of the estimated slope parameter $\mu$ (left) and the corresponding log-decay trends (right) for the Toys4K and OmniObject3D datasets. The red lines indicate the median $\mu$ values, which are obtained via RANSAC fitting on 100 sampled instances and utilized as the data-driven parameters.}
    \label{aaa}
\end{figure}

\begin{table}[t]
\centering
\caption{\textbf{Quantitative results via data-driven parameter selection for two datasets.}}
\label{bbb}
\resizebox{\columnwidth}{!}{
\begin{tabular}{lcccc}
\toprule
\textbf{Dataset} & \textbf{Slope $\mu$} & \textbf{Speed (Iter/s)}$\uparrow$ & \textbf{CD}$\downarrow$ & \textbf{F-Score}$\uparrow$ \\
\midrule
Toys4k & -0.07 (Default) & 0.5850 & \textbf{0.0697} & \textbf{54.0900} \\
&-0.169 (Data-driven) & \textbf{0.6039} &  0.0710 & 53.3538 \\         
\midrule
Omniobject3D & -0.07 (Default) & \textbf{0.6294} & 0.0428  & 74.2281 \\
&-0.183 (Data-driven) & 0.6217 &  \textbf{0.0424} & \textbf{74.5404} \\
\bottomrule
\end{tabular}
}
\end{table}

\subsection{Data-driven Parameter Selection}
Specifically, we randomly sample 100 instances from each dataset and perform RANSAC fitting on the intermediate steps (steps 8 to 18), while filtering out outlier samples. Then we adopt the median slope from this subset. In Fig. \ref{aaa} and Table \ref{bbb}, this data-driven approach yields comparable metrics and even superior performance on two datasets.

\subsection{More Ablation Study} 
More ablation studies are conducted in Phase 3, including the fixed sampling ratio $\xi$ in Table~\ref{xiab} and full sampling step $f_\text{corr}$ in Table~\ref{fab}. Combined with the ablation studies in $\xi$ and $f_\text{corr}$, we determine \textbf{the default hyperparameters of our Fast3Dcache method}: $\mu=-0.07$, $\omega=0.7$, $\tau=3$, $\xi=0.7$, $f_\text{corr}=3$, $\rho_\text{a} = 0.2, \rho_\text{CFG-OFF} = 0.75$. The hyperparameters can be controlled flexibly based on users' requirement of quality and speed trade-off.

\begin{table}[t]
  \centering
  \caption{\textbf{Ablation study of the hyperparameter $\xi$.} CD of $\xi=0.7$ and F-Score of $\xi=0.9$ is better than those of other strategy. In default hyperparameter in our method, $\xi=0.7$ is chosen.}
  \label{xiab}
  \begin{tabular}{lccc}
    \toprule
    \textbf{Strategy}  & \textbf{CD$\downarrow$} & \textbf{F-Score$\uparrow$} \\
    \midrule
    Full sample  & 0.0699 & 54.0608\\
    \midrule
    Aggressive ( $\xi = 0.7$ ) & \textbf{0.0697} & 54.0900 \\
    Aggressive ( $\xi = 0.8$ ) & 0.0698 & 54.0776 \\
    Aggressive ( $\xi = 0.9$ ) & 0.0698 & \textbf{54.1517} \\
    \bottomrule
  \end{tabular}
\end{table}

\begin{table}[t]
  \centering
  \caption{\textbf{Ablation study of the hyperparameter $f_\text{corr}$.} Based on our observations of visualization results, $f_\text{corr}=0$ does not obtain better quality although it gets better metrics of CD and F-Score. And the FLOPs of $f_\text{corr}=0$ is more than that of $f_\text{corr}=3$. After careful consideration of balancing speed and quality, \textbf{we set $f_\text{corr}=3$ as the default parameter of Fast3Dcache.}}
  \label{fab}
  \resizebox{\columnwidth}{!}{
  \begin{tabular}{lcccc}
    \toprule
    \textbf{Strategy}  & \textbf{FLOPs (T)$\downarrow$} & \textbf{CD$\downarrow$} & \textbf{F-Score$\uparrow$} \\
    \midrule
    $f_\text{corr} = 0$ (w.o. $f_\text{corr}$) & \underline{138.0} & \textbf{0.0695} & \textbf{54.1603}\\
    \midrule
    $f_\text{corr} = 1$ (Full sampling) & 155.5 & 0.0700 & 53.9062\\
    $f_\text{corr} = 2$ & 146.7 & 0.0698 & 54.0241\\
    $f_\text{corr} = 3$ & \textbf{115.4} & \underline{0.0697} & \underline{54.0900}\\

    \bottomrule
  \end{tabular}}
\end{table}

\subsection{More Results of Hyperparameter Sensitivity}
We analyze the sensitivity of the model to the slope parameter $\mu$ and the refresh interval $\tau$, as shown in the Table \ref{muandtau}. The results on the left indicate that our method maintains robust performance despite small variations in $\mu$. Regarding $\tau$ (right), it serves as a flexible control for the speed-quality trade-off rather than a sensitive hyperparameter requiring tuning. A larger $\tau$ consistently yields higher acceleration.

\begin{table}[t]
\centering
\caption{\textbf{Hyperparameter analysis of $\mu$ (left) and $\tau$ (right).}}
\label{muandtau}
\resizebox{\columnwidth}{!}{%
\begin{tabular}{lccc|lccc}
\toprule
\textbf{$\mu$} & \textbf{Speed (Iter/s)}$\uparrow$ & \textbf{CD}$\downarrow$ & \textbf{F-Score}$\uparrow$ & \textbf{$\tau$} & \textbf{Speed (Iter/s)}$\uparrow$ & \textbf{CD}$\downarrow$ & \textbf{F-Score}$\uparrow$ \\
\midrule
-0.05 & 0.6085 & 0.0709 & 53.6030       & 1 & 0.4853 & 0.0692 & 54.5439 \\
-0.06 & 0.6063 & 0.0706 & 53.6997       & 2 & 0.5825 & 0.0702 & 53.7980 \\
-0.07 & 0.5850 & 0.0697 & 54.0900       & 3 & 0.5850 & 0.0697 & 54.0900 \\
-0.08 & 0.6128 & 0.0711 & 53.3805       & 4 & 0.6253 & 0.0710 & 53.2729 \\
-0.09 & 0.6135 & 0.0714 & 53.2878       & 5 & 0.6344 & 0.0712 & 53.5003 \\ 
\bottomrule
\end{tabular}
}
\end{table}

\section{More Visualizations}
\subsection{More Visualizations of Voxel Dynamics} 
\label{more ooo}
\begin{figure}[t!]
	\centering
	\begin{subfigure}{0.49\linewidth}
		\centering
		\includegraphics[width=1.0\linewidth]{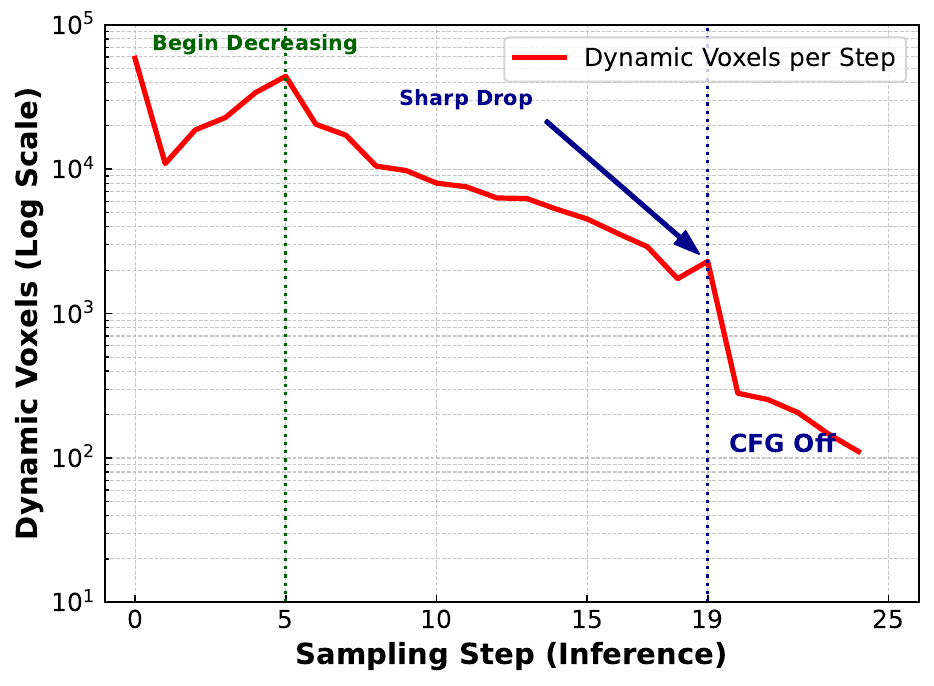}
	\end{subfigure}
    \centering
	\begin{subfigure}{0.49\linewidth}
		\centering
		\includegraphics[width=1.0\linewidth]{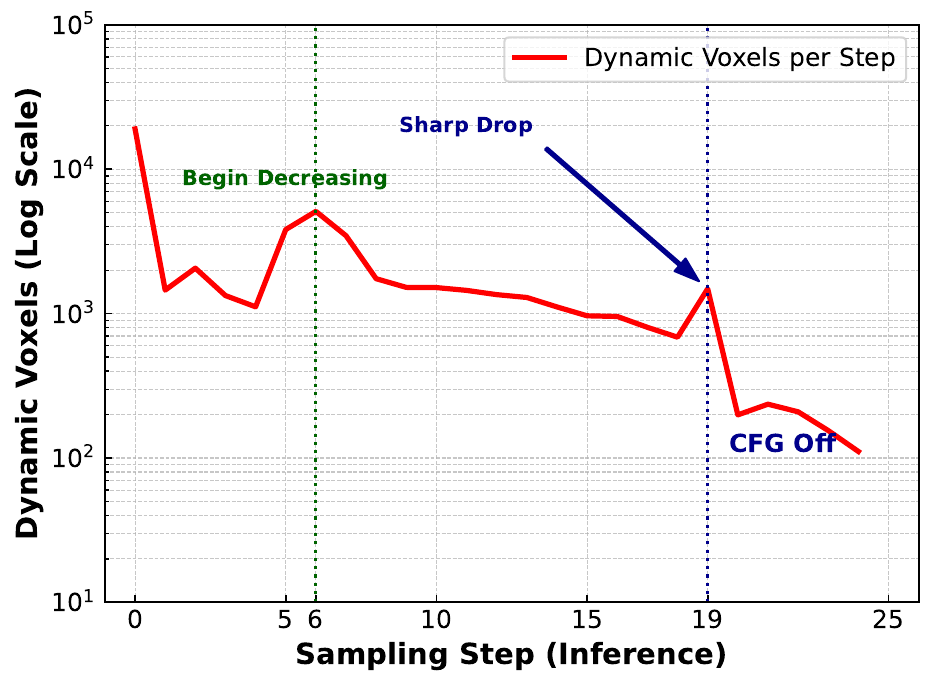}
	\end{subfigure}
    \qquad
	\centering
	\begin{subfigure}{0.49\linewidth}
		\centering
		\includegraphics[width=1.0\linewidth]{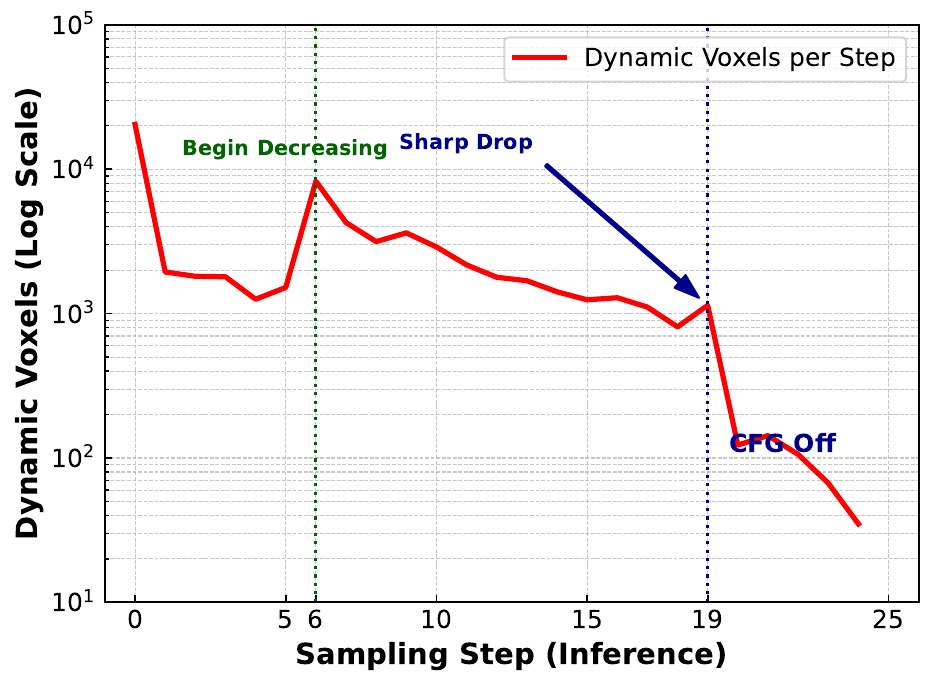}
	\end{subfigure}
    \centering
	\begin{subfigure}{0.49\linewidth}
		\centering
		\includegraphics[width=1.0\linewidth]{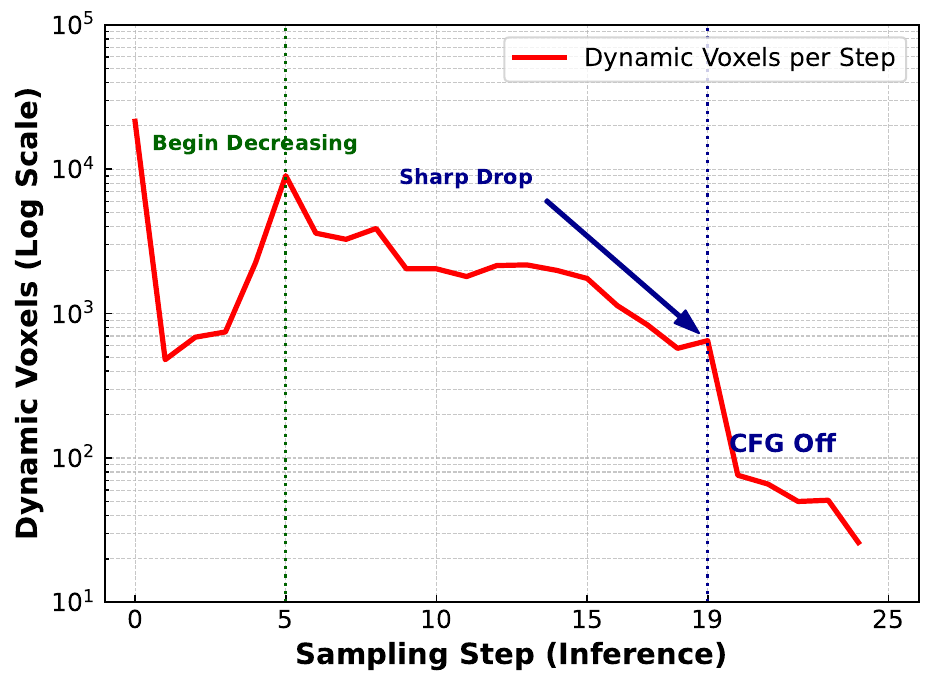}
	\end{subfigure}
    \qquad
	\centering
	\begin{subfigure}{0.49\linewidth}
		\centering
		\includegraphics[width=1.0\linewidth]{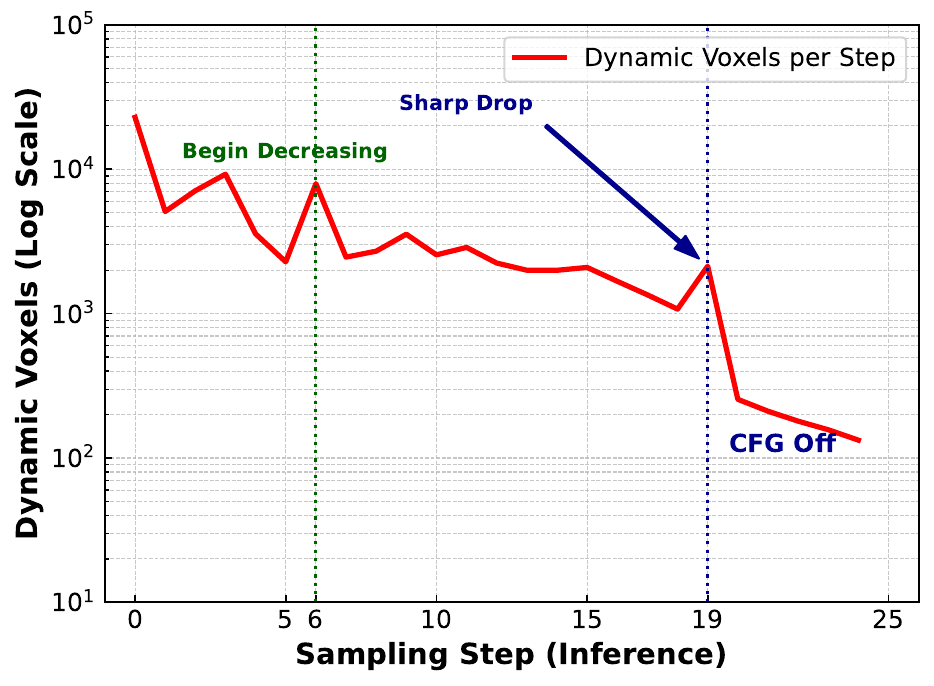}
	\end{subfigure}
    \centering
	\begin{subfigure}{0.49\linewidth}
		\centering
		\includegraphics[width=1.0\linewidth]{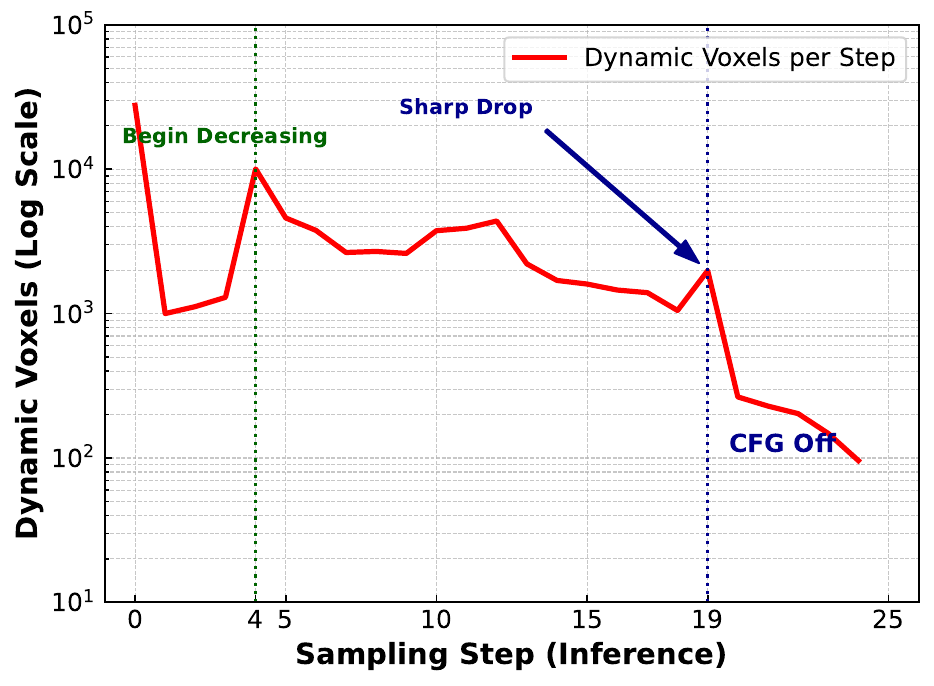}
	\end{subfigure}
    \qquad
	\centering
	\begin{subfigure}{0.49\linewidth}
		\centering
		\includegraphics[width=1.0\linewidth]{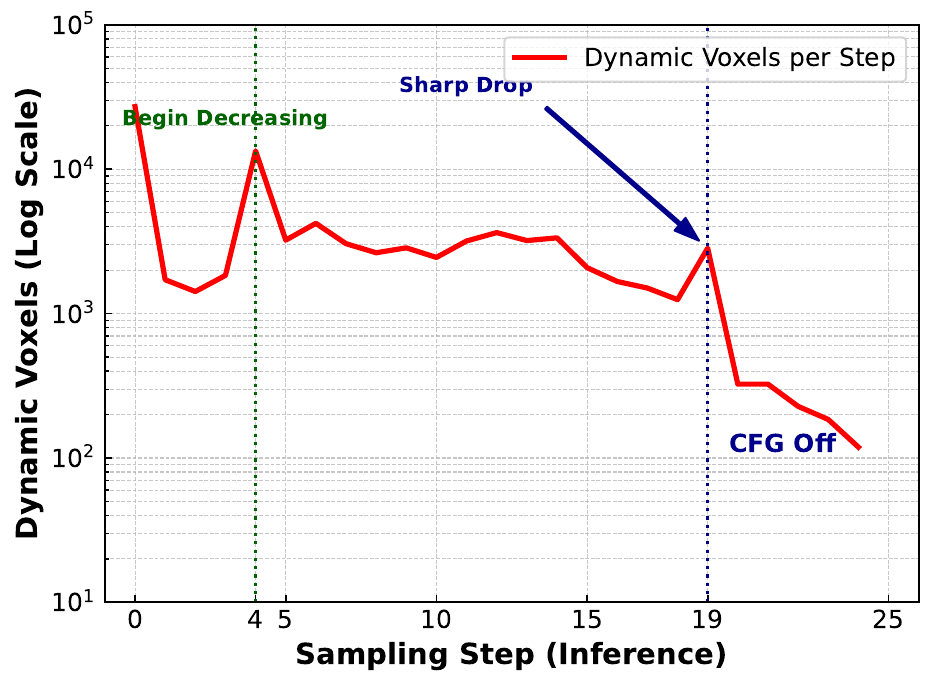}
	\end{subfigure}
    \centering
	\begin{subfigure}{0.49\linewidth}
		\centering
		\includegraphics[width=1.0\linewidth]{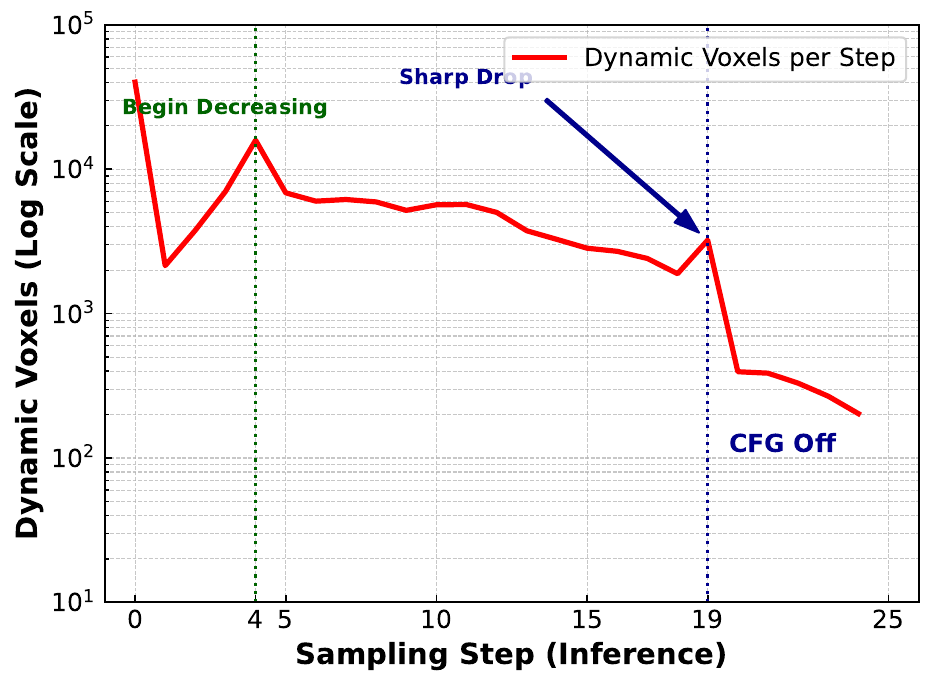}
	\end{subfigure}
	\caption{\textbf{More visualizations of dynamic voxels in inferences of different cases.} Phase 1 is unstable which is implemented that the outline is being formed. In Phase 2, the number of dynamic voxels starts to decrease and can be predicted via PCSC. Despite the fluctuations in the downward trend during the second phase, the experimental results confirm that the log-linear approximation is acceptable. Phase 3 is also CFG Off Phase.}
	\label{varimore}
\end{figure}

To further validate the universality of our PCSC design, we present extended visualizations of voxel evolution across diverse input prompts in Fig.~\ref{varimore}. Consistent with our primary findings, Phase 2 exhibits a stable decay in dynamic voxels across all test cases. These empirical results strongly corroborate the efficacy of using a log-linear approximation to predict the caching budget.

\subsection{More Visualizations of Velocity Field}
\label{more vira}
\begin{figure*}[t!]
	\centering
	\begin{subfigure}{1\linewidth}
		\centering
		\includegraphics[width=0.9\linewidth]{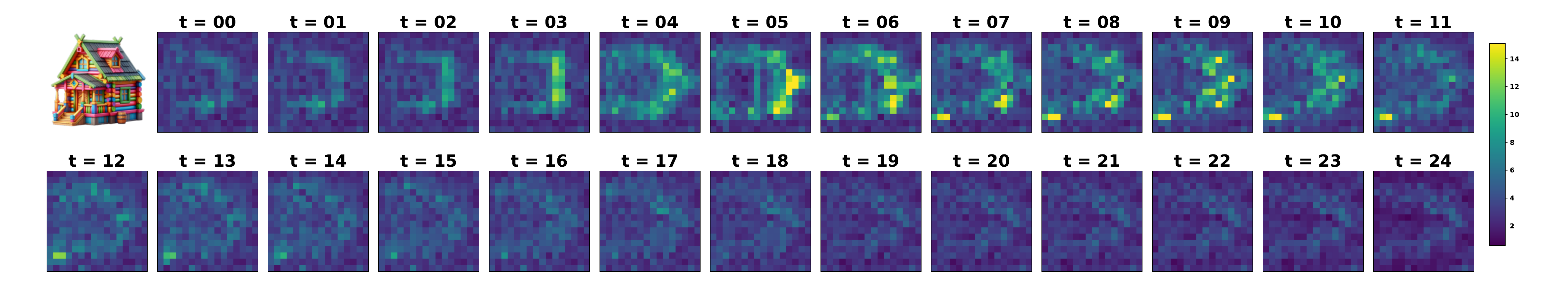}
		\label{v111}
	\end{subfigure}
    \qquad
    \centering
	\begin{subfigure}{1\linewidth}
		\centering
		\includegraphics[width=0.9\linewidth]{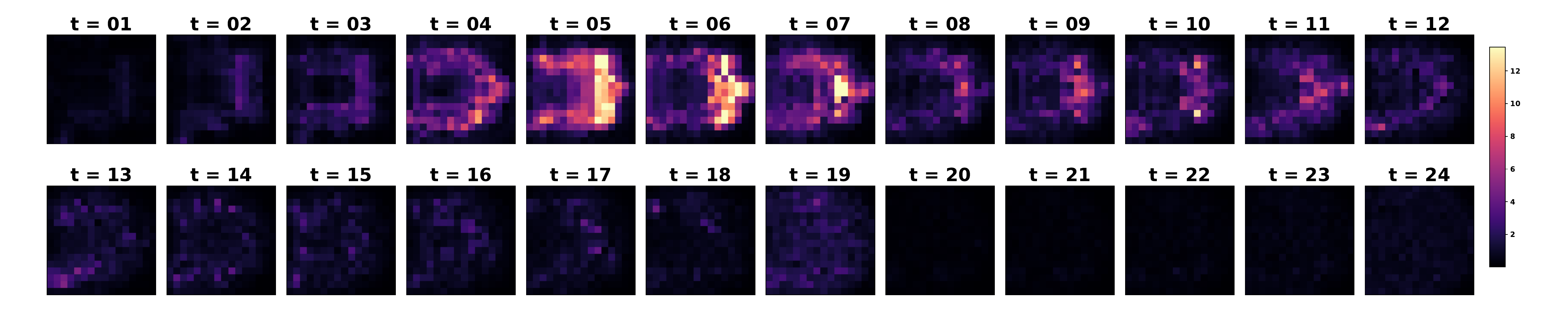}
		\label{a111}
	\end{subfigure}
	\caption{\textbf{Visualization of velocity field and acceleration field feature maps in $\mathbf{\mathcal{S}}_t$ with prompt example 1.}}
	\label{va111111}
\end{figure*}

\begin{figure*}[t!]
	\centering
	\begin{subfigure}{1\linewidth}
		\centering
		\includegraphics[width=0.9\linewidth]{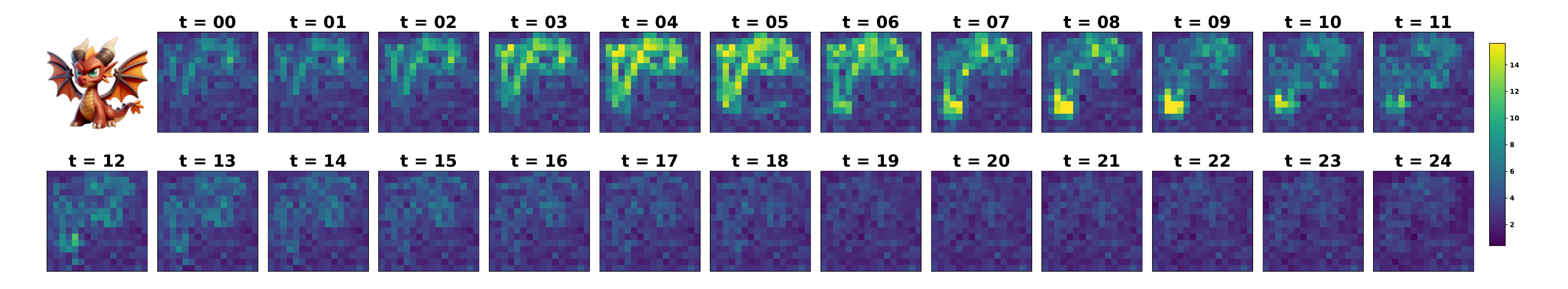}
		\label{v222}
	\end{subfigure}
    \qquad
    \centering
	\begin{subfigure}{1\linewidth}
		\centering
		\includegraphics[width=0.9\linewidth]{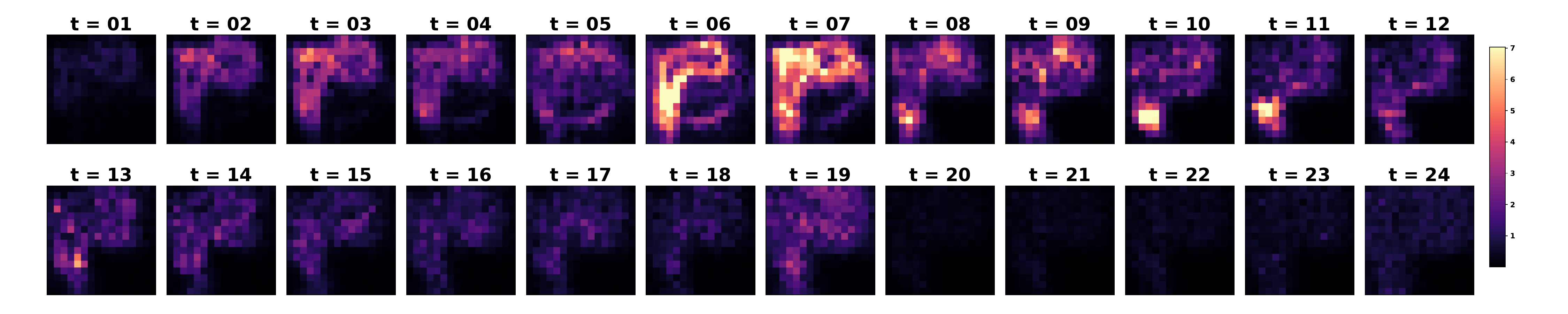}
		\label{a222}
	\end{subfigure}
	\caption{\textbf{Visualization of velocity field and acceleration field feature maps in $\mathbf{\mathcal{S}}_t$ with prompt example 2.}}
	\label{va222222}
\end{figure*}

\begin{figure*}[t!]
	\centering
	\begin{subfigure}{1\linewidth}
		\centering
		\includegraphics[width=0.9\linewidth]{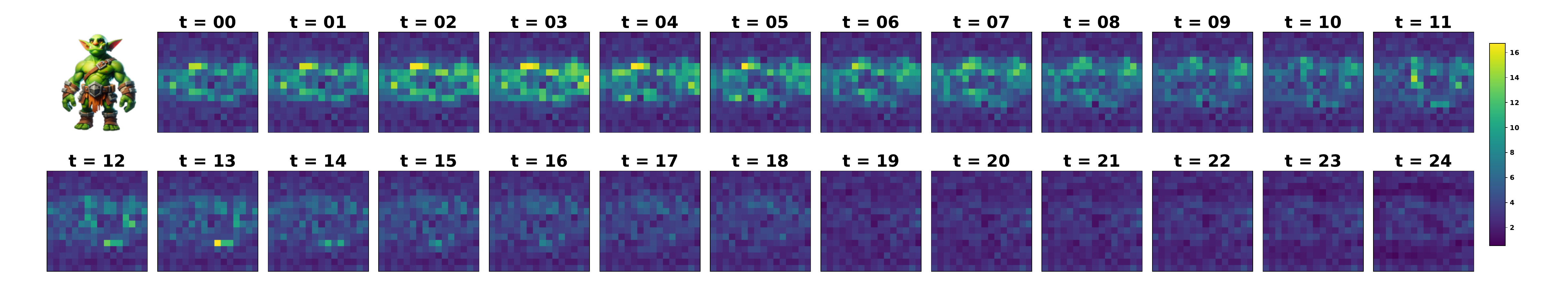}
		\label{v333}
	\end{subfigure}
    \qquad
    \centering
	\begin{subfigure}{1\linewidth}
		\centering
		\includegraphics[width=0.9\linewidth]{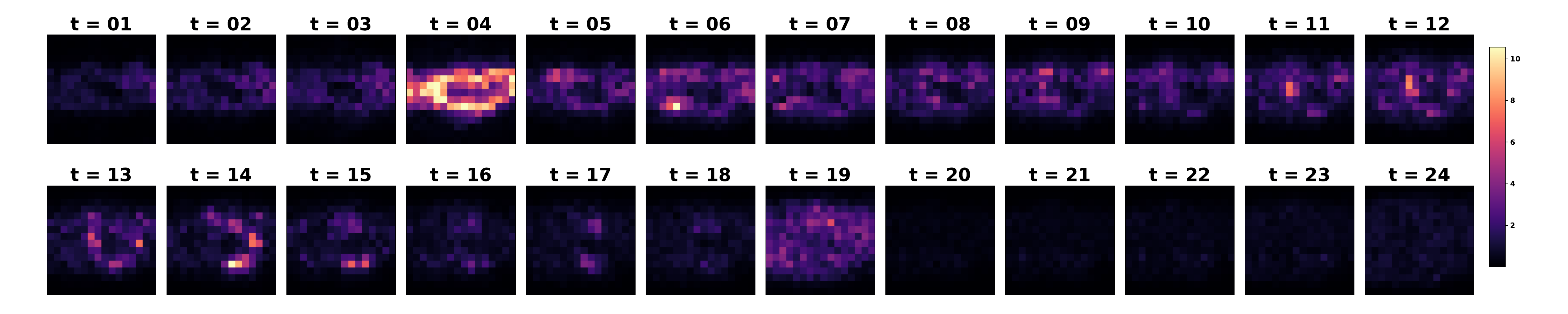}
		\label{a333}
	\end{subfigure}
	\caption{\textbf{Visualization of velocity field and acceleration field feature maps in $\mathbf{\mathcal{S}}_t$ with prompt example 3.}}
	\label{va333333}
\end{figure*}

\begin{figure*}[t!]
	\centering
	\begin{subfigure}{1\linewidth}
		\centering
		\includegraphics[width=1.0\linewidth]{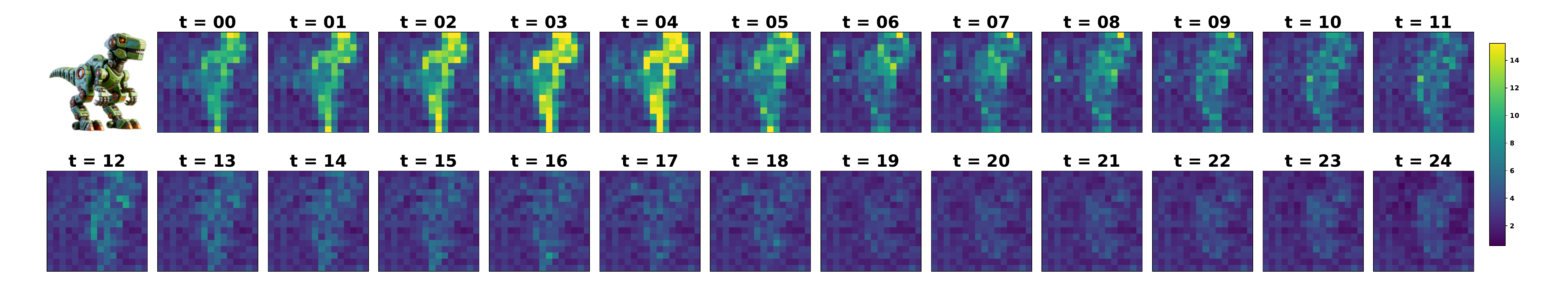}
		\label{v444}
	\end{subfigure}
    \qquad
    \centering
	\begin{subfigure}{1\linewidth}
		\centering
		\includegraphics[width=1.0\linewidth]{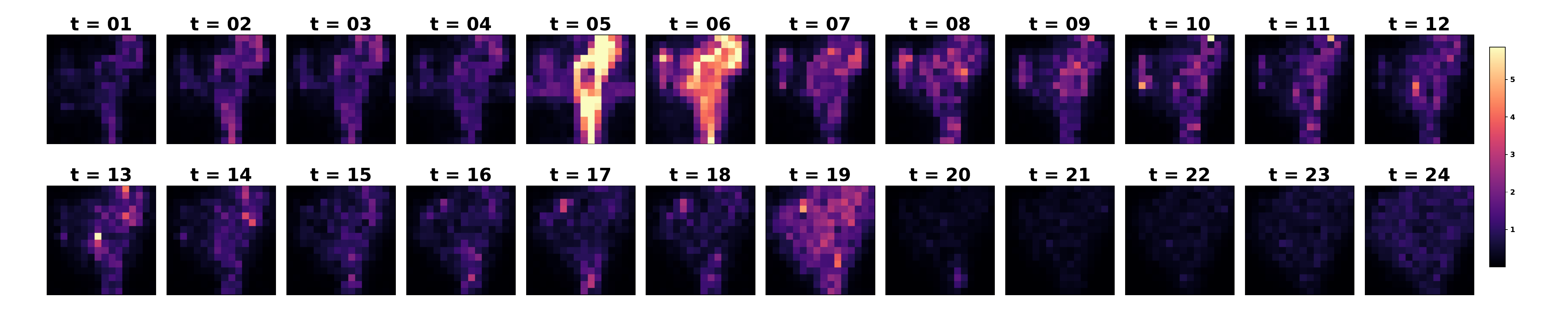}
		\label{a444}
	\end{subfigure}
	\caption{\textbf{Visualization of velocity field and acceleration field feature maps in $\mathbf{\mathcal{S}}_t$ with prompt example 4.}}
	\label{va444444}
\end{figure*}

In Fig.~\ref{va111111} - \ref{va444444}, we visualize more velocity field and acceleration field feature maps in $\mathbf{\mathcal{S}}_t$ to form the observations. The laws are similar with the case in the body of paper and we can leverage these observations to select active voxels during inference.

\subsection{More Visualizations of Generation}
\label{more resultss}
\begin{figure*}[t!]
    \centering
    \includegraphics[width=1.0\linewidth]{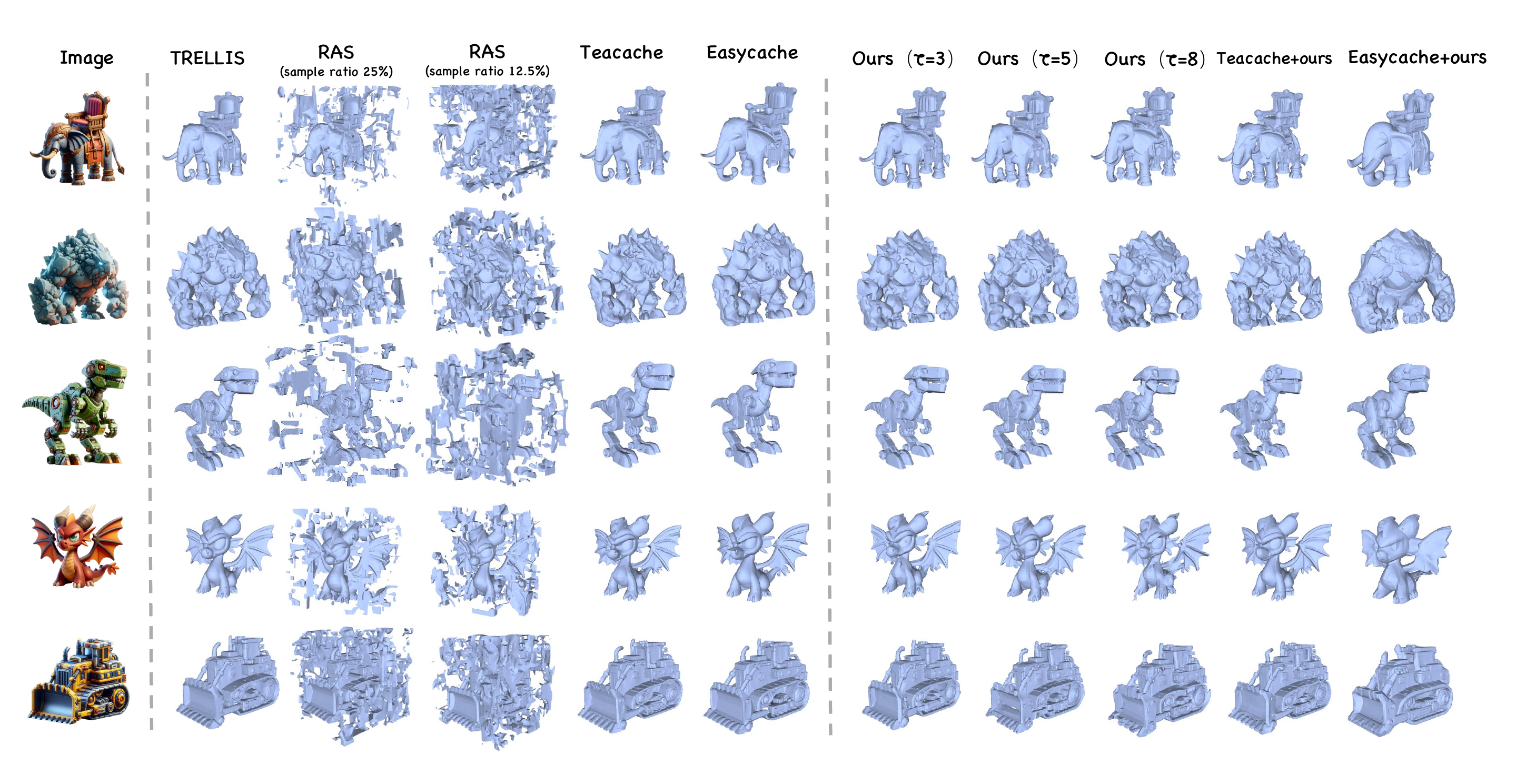}
    \caption{\textbf{More generation visualization results of different methods.}}
    \label{vire1}
\end{figure*}

\begin{figure*}[t!]
    \centering
    \includegraphics[width=1.0\linewidth]{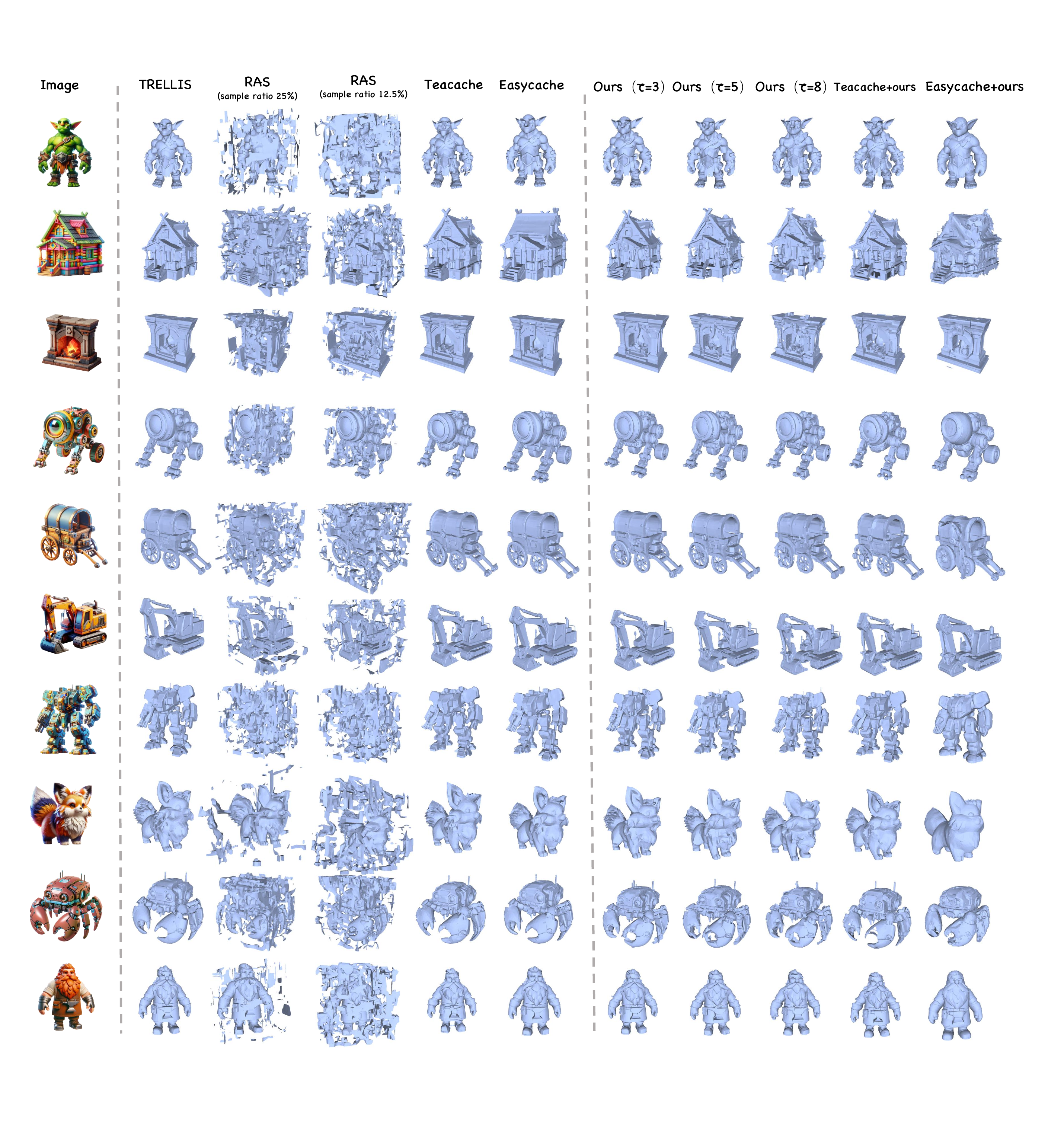}
    \caption{\textbf{More generation visualization results of different methods.}}
    \label{vire2}
\end{figure*}

More visualizations are demonstrated in this section, including Fig.~\ref{vire1} and \ref{vire2}. The prompts are mainly from examples in TRELLIS~\cite{xiang2025structured}, including humans, buildings, normal objects, animals and creative objects. The outcomes of different parameters of $\tau$ obtain the best visualization quality of geometry and our combination methods still maintain a high level of quality.

\section{Impact of Sampling Parameters on Voxel Dynamics} \label{impact}
\begin{figure}[t]
	\centering
	\begin{subfigure}{0.49\linewidth}
		\centering
		\includegraphics[width=1.0\linewidth]{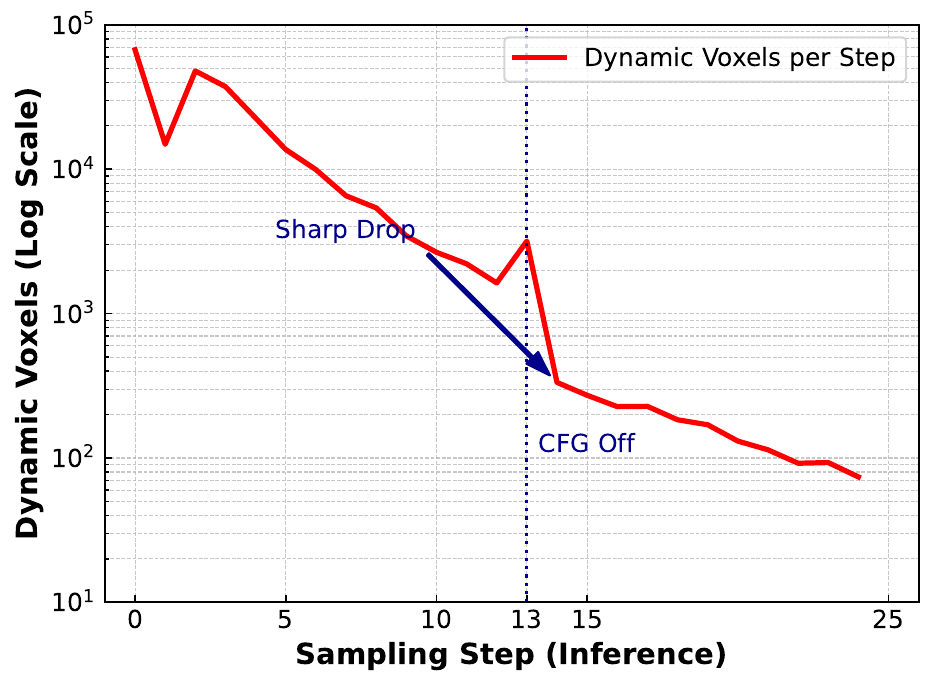}
		\caption{$\eta=1$ with CFG $t \in [0.5,1]$}
        \label{eta1}
	\end{subfigure}
    \centering
	\begin{subfigure}{0.49\linewidth}
		\centering
		\includegraphics[width=1.0\linewidth]{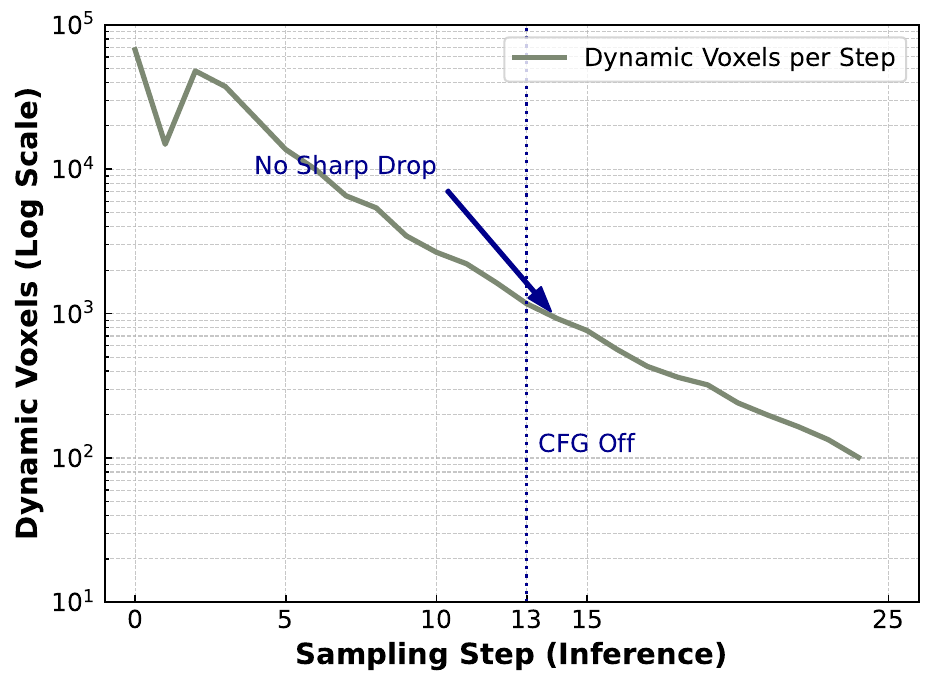}
		\caption{$\eta=1$ with CFG $t \in [0,1]$}
        \label{eta11}
	\end{subfigure}
    \qquad
	\centering
	\begin{subfigure}{0.49\linewidth}
		\centering
		\includegraphics[width=1.0\linewidth]{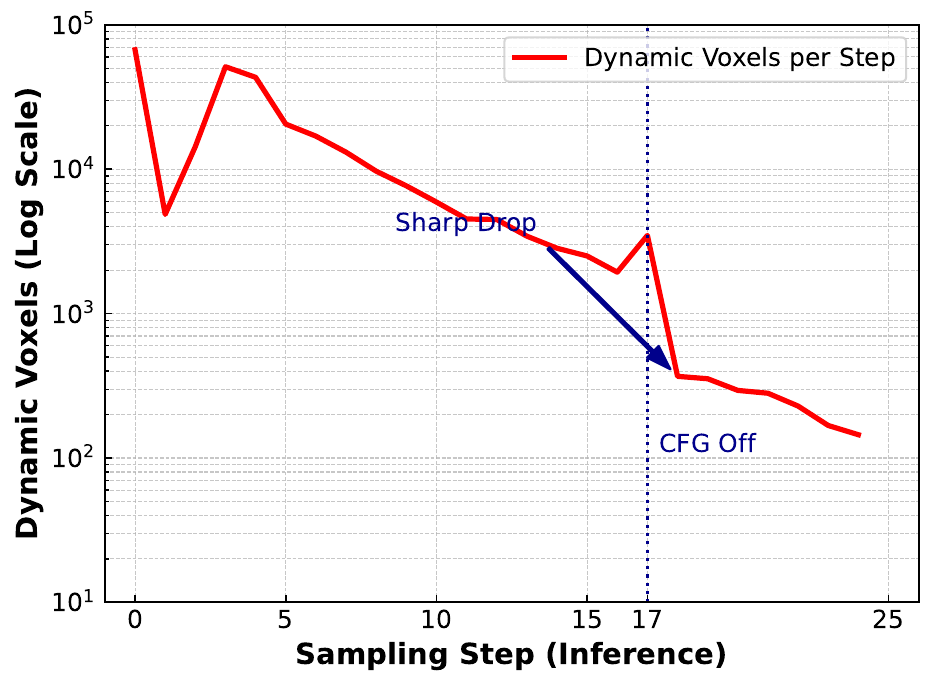}
		\caption{$\eta=2$ with CFG $t \in [0.5,1]$}
        \label{eta2}
	\end{subfigure}
    \centering
	\begin{subfigure}{0.49\linewidth}
		\centering
		\includegraphics[width=1.0\linewidth]{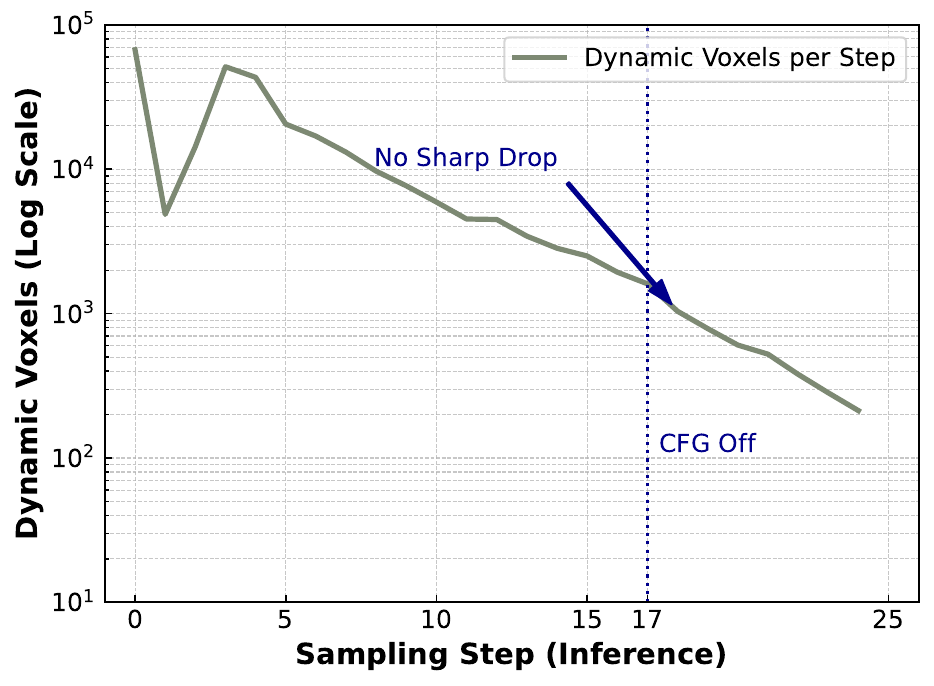}
		\caption{$\eta=2$ with CFG $t \in [0,1]$}
        \label{eta22}
	\end{subfigure}
    \qquad
	\centering
    \begin{subfigure}{0.49\linewidth}
		\centering
		\includegraphics[width=1.0\linewidth]{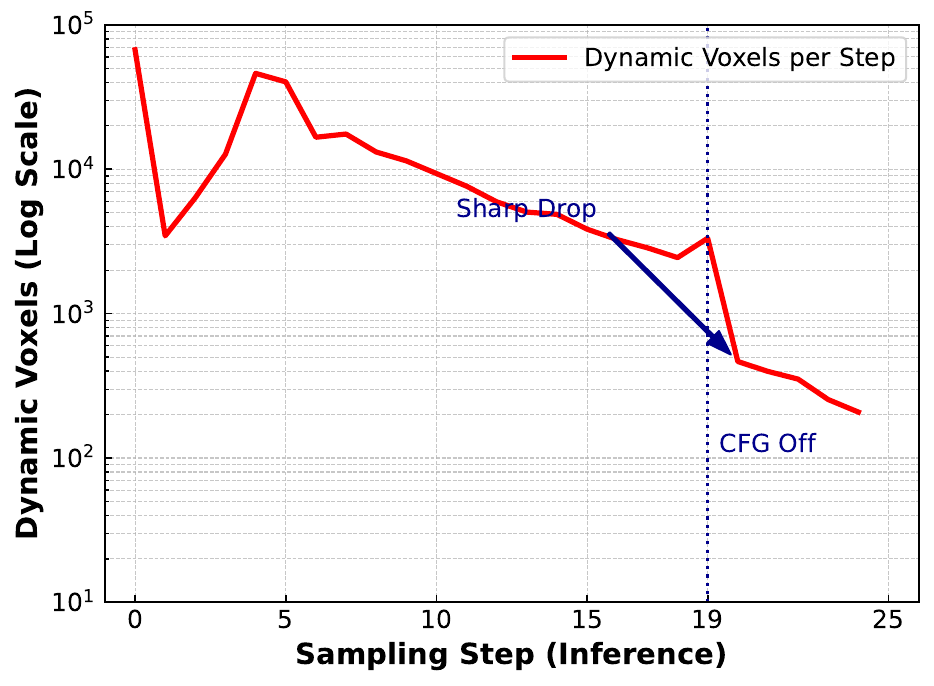}
		\caption{$\eta=3$ with CFG $t \in [0.5,1]$}
        \label{eta3}
	\end{subfigure}
	\centering
    \begin{subfigure}{0.49\linewidth}
		\centering
		\includegraphics[width=1.0\linewidth]{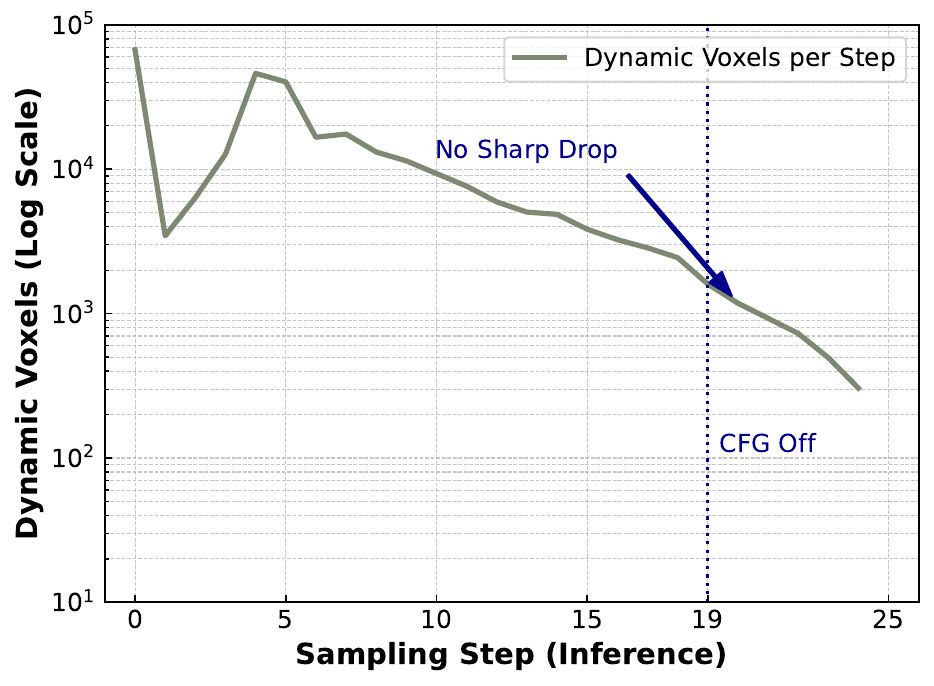}
		\caption{$\eta=3$ with CFG $t \in [0,1]$}
        \label{eta33}
	\end{subfigure}
	\caption{\textbf{The trends of dynamic voxels with different shifting factors $\eta$ and CFG interval.} The point of CFG turning off will result in a significant drop in the number of dynamic voxels. The red plots (a, c, e) correspond to TRELLIS's default CFG interval $t\in[0.5,1]$, where the timing of the "Sharp Drop" is controlled by the shifting factor $\eta$. The green plots (b, d, f) correspond to a full CFG interval $t\in[0,1]$. A direct comparison between the rows (a v.s. b)(c v.s. d)(e v.s. f) demonstrates that continuous CFG guidance removes the sharp drop in dynamic voxels.}
	\label{sup1}
\end{figure}

In Fig.~\ref{sup1}, we investigate the influence of the shifting factor $\eta$ and the Classifier-Free Guidance (CFG) interval on the generation process. Our experiments reveal that voxel stabilization dynamics are highly sensitive to these sampling configurations. Standard Flow Matching implementations typically apply CFG during the interval $t \in [0.5, 1]$ to ensure the initial generation adheres closely to the condition $c$. To optimize this process, a non-uniform time schedule is introduced via the shifting factor $\eta$:
\begin{equation}
t = \frac{\eta \cdot t_\ell}{1 + (\eta - 1) \cdot t_\ell},
\end{equation}
where $t_\ell$ denotes the original timestep in a uniform schedule.

As illustrated in Fig.~\ref{sup1}, setting $\eta > 1$ biases the sampling density, allocating more inference steps to the initial stages governed by CFG. Crucially, the termination of CFG guidance triggers a transition to an unconditional refinement stage, resulting in a precipitous drop in the number of active voxels. The parameter $\eta$ determines the precise step index where this transition occurs. We verify this theoretically and empirically:\begin{itemize}\item Uniform Schedule ($\eta=1$): The transition occurs at the midpoint of the inference process (Fig.~\ref{eta1}). Conversely, applying continuous CFG ($t \in [0, 1]$) eliminates the sharp drop entirely (Fig.~\ref{eta11}).\item Shifted Schedule ($\eta=2$): Solving for the cutoff $t=0.5$ yields $t_\ell = 1/3$. In a 25-step inference process, this shifts the turning point to step $\lceil 25 \times (1 - 1/3) \rceil = 17$, consistent with Fig.~\ref{eta2}.\item Shifted Schedule ($\eta=3$): This setting further delays the refinement stage, as observed in Fig.~\ref{eta3}.\end{itemize}

Consequently, the choice of $\eta$ and the CFG interval directly dictates the duration of the stabilization phases. While our Fast3Dcache phase division is calibrated to the default TRELLIS~\cite{xiang2025structured} parameters ($\eta=3$, CFG $t \in [0.5, 1]$), the framework remains inherently flexible and can be adapted to arbitrary user-defined schedules.

\section{Impact of Computational Cost on Geometry Quality}
\label{implement}
\begin{figure}[t]
    \centering
    \includegraphics[width=1.0\linewidth]{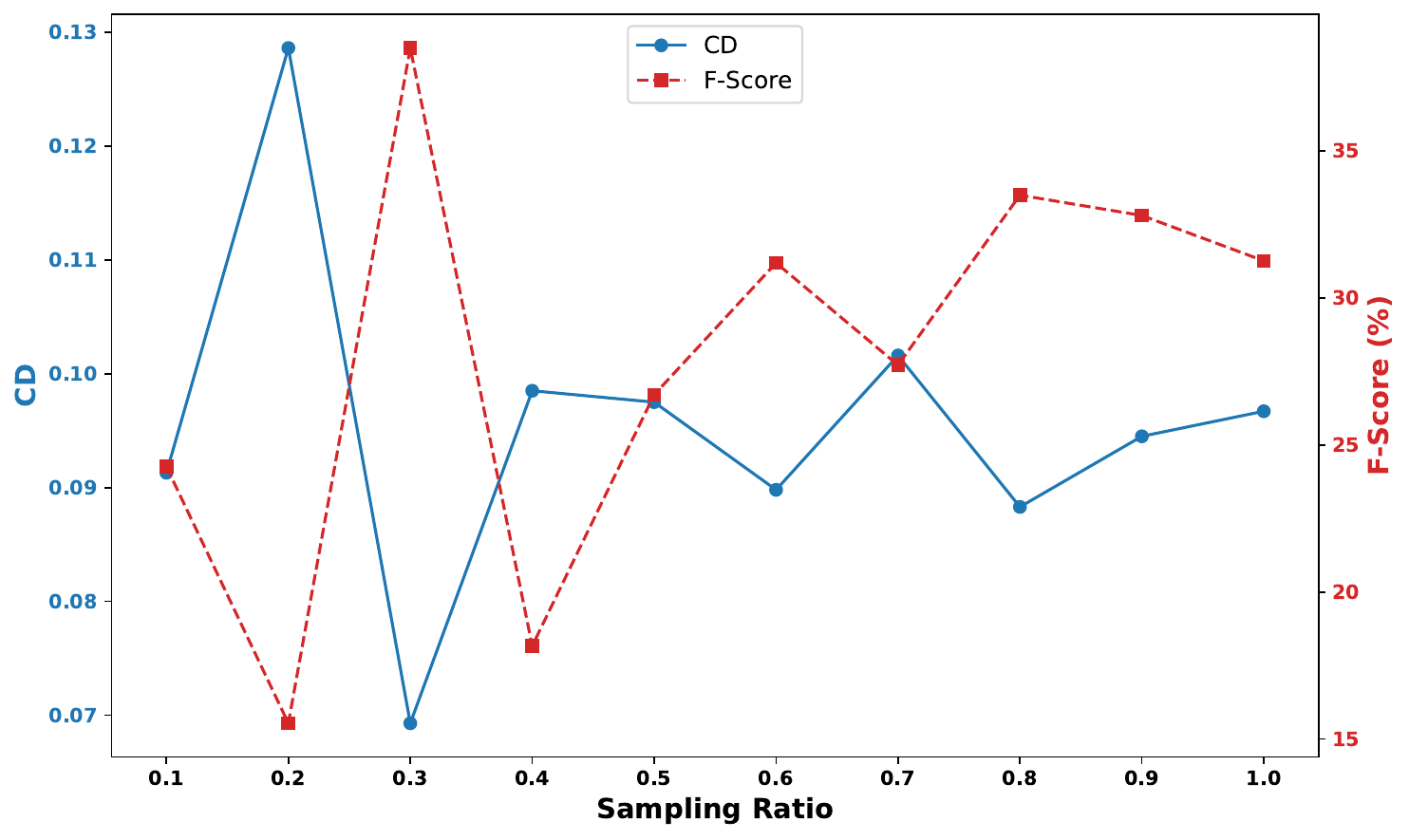}
    \caption{\textbf{Impact of sampling ratio on geometric quality.} The plot illustrates the relationship between sampling ratio (computational volume) and geometric metrics (CD and F-Score). Counter-intuitively, it reveals that a higher computational volume does not consistently lead to superior generation quality. For example, a sampling ratio of 0.3 yields a lower CD (better quality) compared to 0.4, indicating that judicious selection of sampling density perhaps improve fidelity while reducing computation.}
    \label{CDF}
\end{figure}
Fig.~\ref{CDF} illustrates the counter-intuitive phenomenon observed in our quantitative results: higher computational cost does not strictly correlate with superior generation quality. Interestingly, we observe that lower sampling ratios can yield better geometric metrics (lower CD, higher F-Score) in certain regimes. This empirical evidence underscores the feasibility of our approach, demonstrating that it is possible to achieve significant acceleration while simultaneously maintaining or even enhancing geometric fidelity.

\section{Algorithm of Fast3Dcache}
\label{algorithm1}
The details of the Fast3Dcache algorithm are presented in Algorithm~\ref{alg:fast3dcache}.

\begin{algorithm*}[t]
    \caption{Fast3Dcache Inference Pipeline}
    \label{alg:fast3dcache}
    \begin{algorithmic}[1]
        \Require Initial feature grid $\mathcal{S}_\epsilon$, Total steps $T$, Condition $c$
        \Ensure $\mathcal{S}_0$: The final denoised feature grid
        \State \textbf{Hyperparameters:} $\rho_a, \mu, \tau, \omega, \xi, \rho_{\text{CFG-OFF}}, f_{\text{corr}}$
        \Function{SSC}{$\mathbf{v}_{t-1}, \mathbf{v}_{t-2}, c_t, \tau$}
            \State Compute stability score $\mathcal{C}_i(t)$ via velocity and acceleration (weighted by $\omega$).
            \State $\mathcal{I}_{\text{cache}} \gets$ Indices of top-$c_t$ tokens with the \textbf{lowest} $\mathcal{C}_i(t)$
            \State \textbf{if} consecutive caches $\ge \tau$ \textbf{then} $\mathcal{I}_{\text{cache}} \gets \emptyset$ \Comment{\textit{Error Accumulation Elimination}}
            \State $\mathcal{I}_{\text{active}} \gets \text{All Indices} \setminus \mathcal{I}_{\text{cache}}$
            \State \Return $\mathcal{I}_{\text{active}}$
        \EndFunction
        \State Initialize $\mathcal{S}_t \gets \mathcal{S}_\epsilon$, $\mathbf{v}_{\text{cache}} \gets \mathbf{0}$
        \State Pre-calculate cache budget schedule $N_{\text{cache}}(t)$ using PCSC curve.
        \For{$k \gets 1$ to $T$}
            \State $t \gets t_k$, $t_{\text{prev}} \gets t_{k+1}$
            \Statex \Comment{\textit{Step 1: Determine Cache Budget $c_t$}}
            \If{$k \le \lceil T \cdot \rho_\text{a} \rceil$} \Comment{Phase 1: Full Sampling}
                \State $c_t \gets 0$
            \ElsIf{$k < \lceil T \cdot \rho_{\text{CFG-OFF}} \rceil$} \Comment{Phase 2: Dynamic Caching}
                \State $c_t \gets N_{\text{cache}}(t)$
            \Else \Comment{Phase 3: CFG-Free Refinement}
                \State $k_{\text{refine}} \gets k - \lceil T \cdot \rho_{\text{CFG-OFF}} \rceil$
                \If{$(k_{\text{refine}} + 1) \pmod{f_{\text{corr}}} = 0$}
                    \State $c_t \gets 0$ \Comment{Full correction step}
                \Else
                    \State $c_t \gets D^3 \cdot \xi$ \Comment{Fixed ratio caching}
                \EndIf
            \EndIf
            
            \Statex \Comment{\textit{Step 2: Token Selection \& Model Inference}}
            \State $\mathcal{I}_{\text{active}}^{(t)} \gets \Call{SSC}{\mathbf{v}_{\text{cache}}, \mathbf{v}_{\text{prev\_cache}}, c_t, \tau}$
            \State $\mathbf{v}_{\text{active}} \gets \operatorname{FlowTransformer}(\mathcal{S}_t[\mathcal{I}_{\text{active}}^{(t)}], t, c)$
            
            \Statex \Comment{\textit{Step 3: State Update}}
            \State $\mathbf{v}_t \gets \mathbf{v}_{\text{cache}}$
            \State $\mathbf{v}_t[\mathcal{I}_{\text{active}}^{(t)}] \gets \mathbf{v}_{\text{active}}$ \Comment{Update active tokens, reuse others}
            \State $\mathcal{S}_t \gets \mathcal{S}_t - (t - t_{\text{prev}}) \cdot \mathbf{v}_t$
            \State $\mathbf{v}_{\text{prev\_cache}} \gets \mathbf{v}_{\text{cache}}$; $\mathbf{v}_{\text{cache}} \gets \mathbf{v}_t$
        \EndFor
        \State \Return $\mathcal{S}_0$
    \end{algorithmic}
\end{algorithm*}

\section{FLOPs Calculation}
\label{Flops}
For the metric FLOPs, we mainly calculate the floating point operations inside flow transformer blocks because the computational workload of other modules is significantly less than that within the flow transformer.
\begin{enumerate}
  \item Modulation (conditional):
\begin{align*}
\text{FLOPs}_\text{Mod-Block} \approx {} & 5 \times B \times D_\text{model} \\
& + 2 \times B \times D_\text{model} \times (6 D_\text{model}).
\end{align*}
\par
\item LayerNorm 1:
$$\text{FLOPs}_\text{LN-Block} \approx 7 \times B \times N_\text{tok} \times D_\text{model}.$$ \par
\item Self-Attention:
\begin{align*}
\text{FLOPs}_\text{SA} \approx {} & \underbrace{2 B N_\text{tok} D_\text{model} (3 D_\text{model})}_{\text{QKV}} + \\
& \underbrace{2 B H N_\text{tok}^2 (D_\text{model}/H)}_{\text{QK}^T} +  
\underbrace{5 B H N_\text{tok}^2}_{\text{Softmax}} \\
& + \underbrace{2 B H N_\text{tok}^2 (D_\text{model}/H)}_{\text{Attn $\times$ V}} + \\
& \underbrace{2 B N_\text{tok} D_\text{model}^2}_{\text{OutProj}},
\end{align*}
so, $\text{FLOPs}_\text{SA} \approx 8 B N_\text{tok} D_\text{model}^2 + 4 B N_\text{tok}^2 D_\text{model} + 5 B H N_\text{tok}^2.$\par
\item LayerNorm 2 is the same as LayerNorm 1.\par
\item Cross-Attention:
\begin{align*}
\text{FLOPs}_{CA} \approx {} & \underbrace{2 B N_\text{tok} D_\text{model}^2}_{\text{Q}} + \underbrace{2 B N_\text{cond} D_\text{cond} (2 D_\text{model})}_{\text{KV}} + \\
& \underbrace{2 B H N_\text{tok} N_\text{cond} (D_\text{model}/H)}_{\text{QK}^T}  \\
 & + \underbrace{5 B H N_\text{tok} N_\text{cond}}_{\text{Softmax}} + \\
& \underbrace{2 B H N_\text{tok} N_\text{cond} (D_\text{model}/H)}_{\text{Attn $\times$ V}} + \\
& \underbrace{2 B N_\text{tok} D_\text{model}^2}_{\text{OutProj}},
\end{align*}
so, $\text{FLOPs}_\text{CA} \approx 4 B N_\text{tok} D_\text{model}^2 + 4 B N_\text{cond} D_\text{model}^2 + 4 B N_\text{tok} N_\text{cond} D_\text{model} + 5 B H N_\text{tok} N_\text{cond}.$\par
\item LayerNorm 3 is the same as LayerNorm 1.\par
\item FFN (MLP):
\begin{align*}
\text{FLOPs}_\text{MLP} \approx {} & \underbrace{2 B N_\text{tok} D_\text{model} D_\text{mlp}}_{\text{fc1}} + \underbrace{5 B N_\text{tok} D_\text{mlp}}_{\text{Activation}} \\
& + \underbrace{2 B N_\text{tok} D_\text{mlp} D_\text{model}}_{\text{fc2}},
\end{align*}
because $R_\text{mlp}=4, D_\text{mlp}=4D_\text{model}$,
$$\text{FLOPs}_\text{MLP} \approx 16 B N_\text{tok} D_\text{model}^2 + 20 B N_\text{tok} D_\text{model}.$$\par
\end{enumerate}
Overall,
\begin{align*}
\text{FLOPs}_\text{Block} = {} & \text{FLOPs}_\text{Mod-Block} + 3 \times \text{FLOPs}_\text{LN-Block} \\
& + \text{FLOPs}_\text{SA} + \text{FLOPs}_\text{CA} \\
& + \text{FLOPs}_\text{MLP}.
\end{align*}
If the Fast3Dcache method is leveraged, then $N_\text{tok}(t) = c_t = N_\text{active}(t)$.

\section{Limitation and Future Work}
\label{limitation}
Our current implementation of Fast3Dcache is optimized for the sparse voxel grid representation employed by the state-of-the-art TRELLIS~\cite{xiang2025structured} framework. While the core principle of leveraging spatiotemporal redundancy is universally applicable, applying our specific voxel-based metrics to continuous or implicit representations (\textit{e.g.} Signed Distance Fields) requires tailoring the stability criteria to those respective domains. In future work, we plan to extend this geometry-aware caching paradigm to a broader spectrum of 3D representations, aiming to establish a unified efficient synthesis framework across diverse modalities. The second stage (SLAT) can leverage established 2D acceleration methods and we plan to address texture-specific constraints in a unified framework in future research.

